\documentclass[10pt]{article}

\usepackage{amsmath, amsfonts, amssymb,  epsfig, mathtools, caption}
\usepackage{times, algorithm, algorithmic, amsbsy,bm,amsthm}
\usepackage{xcolor}
\usepackage[capitalize]{cleveref}
\usepackage{subcaption}
\usepackage{amsmath}
\usepackage{graphicx}

\usepackage[preprint]{neurips_2020}

\newtheorem{theorem}{Theorem}

\newtheorem{definition}{Definition}
\newtheorem{corollary}{Corollary}
\newtheorem{assumption}{Assumption}
\newtheorem{remark}{Remark}

\DeclareMathOperator*{\softmax}{softmax}
\DeclareMathOperator*{\crossentropy}{CrossEntropy}
\usepackage{titling}

\graphicspath{{figures/}}

\def\1{\mathbf 1}

\title{Stragglers Are Not Disaster: A Hybrid Federated Learning Algorithm with Delayed Gradients}
\begin{document}
\maketitle

\begin{abstract}
	Federated learning (FL) is a new machine learning framework which trains a joint model across a large amount of decentralized computing devices. Existing methods, e.g., Federated Averaging (FedAvg), are able to provide an optimization guarantee by synchronously training the joint model, but usually suffer from stragglers, i.e., IoT devices with low computing power or communication bandwidth, especially on heterogeneous optimization problems. To mitigate the influence of stragglers, this paper presents a novel FL algorithm, namely Hybrid Federated Learning (HFL), to achieve a learning balance in efficiency and effectiveness. It consists of two major components: synchronous kernel and asynchronous updater. Unlike traditional synchronous FL methods, our HFL introduces the asynchronous updater which actively pulls unsynchronized and delayed local weights from stragglers. An adaptive approximation method, Adaptive Delayed-SGD (AD-SGD), is proposed to merge the delayed local updates into the joint model.
	The theoretical analysis of HFL shows that the convergence rate of the proposed algorithm is $\mathcal{O}(\frac{1}{t+\tau})$ for both convex and non-convex optimization problems. 
\end{abstract}

\section{Introduction}	\label{Sec:Introduction}
Federated Learning (FL) \cite{konevcny2016federated}, has emerged as an attractive paradigm for training a joint model in a federated network. Compared to the standard parallel optimization framework where the model is trained with the large-scaled dataset on a central server \cite{shalev2014understanding, goodfellow2016deep}, FL trains a joint model under the coordination of a server across a large number remote devices. The joint model is learned with the updates from the remote devices via local training on their private data \cite{johansson2007simple, lee2013distributed, bonawitz2017practical, mcmahan2017communication}. 
Since the model is trained without sharing data, the data privacy can be greatly enhanced in FL. 

Two key challenges arise in FL. First, FL typically trains the model on a \textit{heterogeneous network}, where the remote devices are large in number and have a variety of computing power and communication bandwidth. This makes the training process suffer from many stragglers which are slow in their local model training. 
Second, FL usually trains on a \textit{heterogeneous dataset}, where training data are highly unbalanced and non-i.i.d. 
To tackle these challenges, several FL frameworks have been studied in literature. For example, Federated Averaging (FedAvg)  \cite{mcmahan2017communication} is developed to address the communication constraint by performing multiple local learning steps on a subset of remote devices before uploading the model updates into the server. 

The convergence of FedAvg can be guaranteed \cite{li2019convergence} when the following two assumptions are made: (i) during the learning process, all remote devices are active, and (ii) the server can access all remote devices with equal probabilities. Later works following the first assumption \cite{zhou2017convergence, li2019convergence, woodworth2018graph, wang2019adaptive, yu2019parallel} and following these two assumptions \cite{stich2018local, li2019convergence, khaled2020tighter, karimireddy2020scaffold, qu2020federated, yang2021achieving} have been conducted to further improve FL performance. However, these two assumptions usually do not hold in practical FL applications, due to the fact that straggles would commonly exist in the network. Once the straggler has the same possibility to be chosen by the server, the learning speed will be greatly reduced. Moreover, the data heterogeneity of FL tells that these stragglers could not be excluded from the training process, as they may contain unique local data that could not be found in other devices.

In this paper, we propose a new FL algorithm, called Hybrid Federated Learning (HFL), to enhance the learning performance of FL with the presence of stragglers in the network. The proposed HFL has two key components: a synchronous kernel and an asynchronous updater for two different communication scenarios. The synchronous kernel aims to synchronize those devices (i.e., non-stragglers) that have high enough computation and communication capacities, and performs the same training strategy as FedAvg which synchronously communicates with the selected devices in every communication round. The asynchronous updater aims to incorporate model updates of those stragglers, which could be several steps behind the synchronous kernel, into the joint model training process. 
Moreover, to bridge the gap between the delayed gradients and the optimal gradients from stragglers, we develop an adaptive approximation method called Adaptive Delayed-SGD (AD-SGD). In particular, the proposed AD-SGD first applies the Taylor expansion to approximate the optimal gradient of a straggler from its delayed gradients in a distributed network.  
Meanwhile, an adaptive hyper-parameter controlling mechanism is developed to reduce the bias of Taylor series approximation. 

We evaluate our proposed HFL algorithm through both theoretical analysis and comprehensive experiments. On both convex and non-convex optimization problems, we provide the theoretical convergence guarantee with non-i.i.d distributed data. The discussion of the convergence rate of HFL is also provided. 
The experimental results show that HFL outperforms existing FL algorithms. In summary, the contribution of this paper are as follows
\begin{itemize}
	\item To address the learning of heterogeneous data in FL network with stragglers, we propose a hybrid structured algorithm HFL consisting of a synchronous kernel and an asynchronous updater to jointly train the model, enabling a learning balance in efficiency and effectiveness. 
	\item To obtain the optimal joint model, we develop  an adaptive approximation method AD-SGD to bridge the gap between the delayed gradients and the optimal gradients. 
	\item We show the performance of theoretical analysis and experiments, which guarantees a convergence on both convex and non-convex optimization problems. 
\end{itemize}

\noindent\textbf{Paper organization.} In Section~\ref{Sec:Formulation}, we describe the background of HFL. In Section~\ref{Sec:HFL}, we detail our proposed HFL algorithm. In Section~\ref{Sec:Convergence}, we provide  theoretical analysis on the convergence rate of HFL. In Section~\ref{Sec:Experiments}, we provide experimental results and analysis by comparing HFL with existing algorithms. For the supplementary material, we introduce the related works in Section~A, the detailed demonstration of convergence results in Section~B and the extended experiment settings and results in Section~C.

\section{Background} \label{Sec:Formulation}
\subsection{FL Objective}\label{Subsec:Objective}
\begin{table}[t!]
\centering
    \caption{Notations Summary}
    \scalebox{0.9}{
    \begin{tabular}{ c c } 
     \hline\hline
        $N, i$ &  total number, index of the remote device \\ 
        $F(\cdot), F_i(\cdot)$ & joint objective, local objective of FL \\
        $\mathbf{X}, \mathbf{X}^i$ & total, local learning dataset\\
        $T, t$ &  number, index of global communication rounds \\ 
        $E, e$ &  number, index of local epoch steps \\ 
        $\mathbf{w}_{t}, n$ & joint model after round $t$ and its dimension \\ 
        $\mathbf{w}_{t}^{i}, g (\mathbf{w}_{t}^{i})$ & model, gradient of $i$-th device at round $t$ \\
     \hline
    \end{tabular}}
	\label{Table:notation}
\end{table}
Consider a distributed network which has one central server and $N$ remote devices. Each remote device owns its local private dataset $\mathbf{X}^i$, for $i=1,2,\cdots,N$. We denote the whole training dataset as $\mathcal{X} = \{\mathbf{X}^1, \mathbf{X}^2, \cdots, \mathbf{X}^N\}$. In this paper, we consider the local data are non-i.i.d., i.e., the data distributions for any two remote devices can be different. A FL process starts with an initialization of a model in the server, and iteratively performs local model training in individual remote devices and joint model updating in the server. Table~\ref{Table:notation} summarizes the use of mathematical symbols in this paper, and the learning objective of FL can be formalized as follows

\begin{equation}\label{Eq:Formulation}
\min_{\mathbf{w}} \left\{F(\mathbf{w}) \triangleq \sum_{i=1}^{N} p_i F_i(\mathbf{w}) \right\},
\end{equation}
where $\mathbf{w}$ is the joint model parameter vector, $F_i$ is the local objective function and $p_i$ is the weighted factor of the $i$-th device, where $p_i \geq 0$ and $\sum_{i=1}^{N} p_i = 1$. Specifically, we denote the training samples in $\mathbf{X}^i$ as $\mathbf{X}^i =  \{x_{i,1}, x_{i,2}, \dots, x_{i,n_i} \}$, then the local objective $F_i(\cdot)$ can be defined as follows
\begin{equation}\label{Eq:Local_objective}
\centering
F_i(\mathbf{w}) \triangleq \frac{1}{n_i} \sum_{j=1}^{n_i} \mathcal{L}(\mathbf{w}; x_{i,j}),
\end{equation}
where $\mathcal{L}(\cdot;\cdot)$ is the loss function for all remote devices. 
In each round, the selected remote device receives the current joint model from the server and performs its local training via Stochastic Gradient Descend (SGD) as follows
\begin{equation}\label{Eq:local}
\mathbf{w}_{t+1}^i = \mathbf{w}_{t}^i - \sum_{e=0}^{E-1} \eta_{t} \nabla F_i (\mathbf{w}_{t,e}^i, \mathbf{X}_{t,e}^{i} ),
\end{equation}
where $\mathbf{w}_{t}^{i} \in R^{n}$ is the received joint model from the server at $t$-th round, and $e = 0, 1, \dots, E-1$ represents the local training epoch index and $\eta_t$ is the learning rate at the $t$-th communication round, determined by the server. The second term in Equation~\eqref{Eq:local} denotes the overall model update or gradients from the $i$-th device, denoted by $g (\mathbf{w}_{t}^{i})$. Then, we have $\mathbf{w}_{t+1}^i = \mathbf{w}_{t}^i - \eta_{t} g (\mathbf{w}_{t}^{i})$. After this, the model update $\mathbf{w}_{t+1}^i$ can be sent back to the server to update the joint model through model updating.

\begin{figure*}[t]
    \centering
    \includegraphics[height=0.35\textwidth, width=0.9\columnwidth]{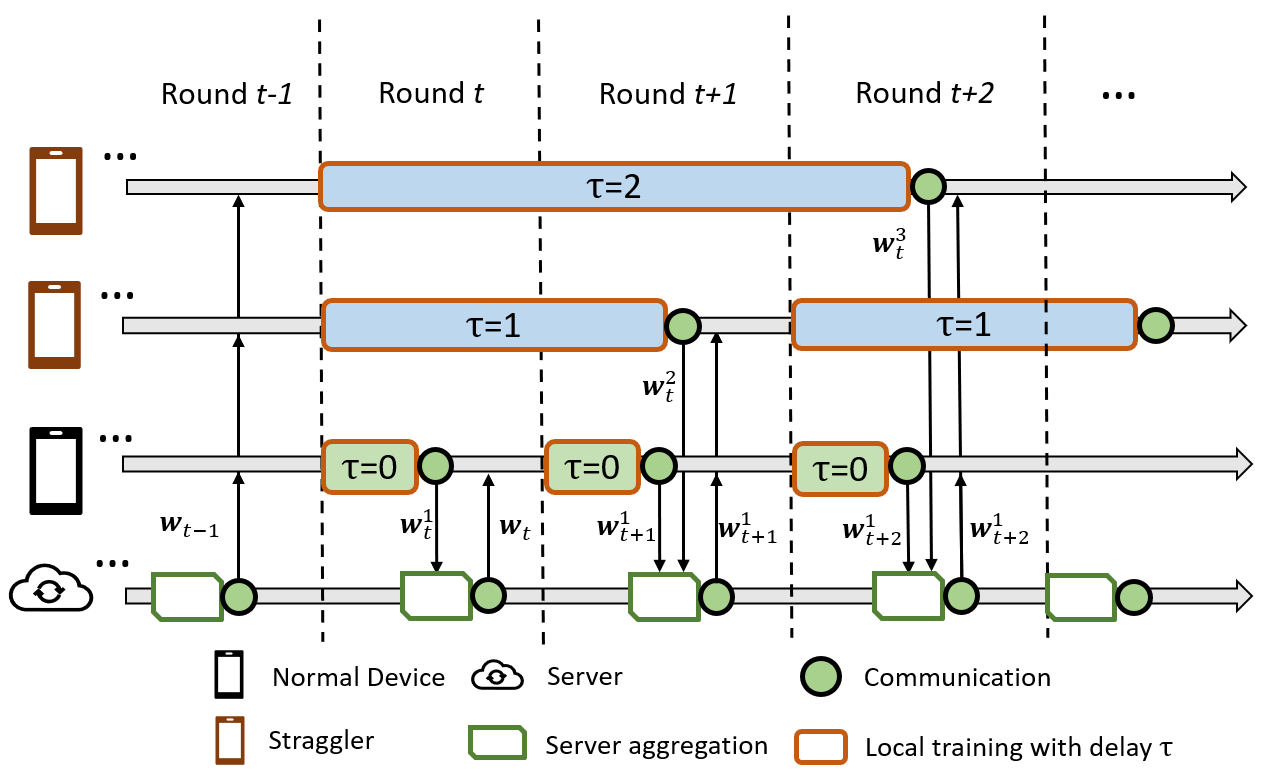}
    \caption{The joint learning process of the proposed HFL using both synchronous kernel and asynchronous updater with the presence of stragglers. One normal device and two stragglers are only illustrated as an example.  }
    \label{Fig:structure}
\end{figure*}

\subsection{Design Motivation}\label{Subsec:Challenge}
It is desired to have all remote devices participating the training process, in order to achieve the optimal performance in solving Equation~\eqref{Eq:Formulation}. However, the presence of stragglers in the distributed network will significantly reduce the training speed of existing synchronous FL methods which is actually determined by the slowest device. 

To reduce the impact of stragglers, some prior studies \cite{stich2018local, li2019convergence, karimireddy2020scaffold, yang2021achieving} have been taken and provide a partial participation algorithm: a threshold $K (1 \leq K \leq N)$ is empirically selected so that the server only accepts model updates from the first $K$ remote devices while discarding the rest $N-K$ slow participators regarded as stragglers. By doing so, the bottleneck of stragglers can be avoided by simply excluding them from the training process. However, due to the data heterogeneity of FL, the removal of those stragglers can greatly reduce the learning performance. Hence, the presence of stragglers in FL usually causes the dilemma of`` efficiency-effectiveness" in learning.

To achieve a balance in learning efficiency and effectiveness, it is necessary to incorporate model updates from stragglers without incurring a long waiting time for synchronization in the server. This motivates us to develop a hybrid learning framework in which both synchronous learning for normal devices and asynchronous learning for slow devices (i.e., stragglers) are considered. However, the incorporation of the delayed model updates from stragglers into the joint model in the server should be carefully designed, which is detailed in the next section.

\section{Hybrid Federated Learning}\label{Sec:HFL}

\subsection{Outline}
The server in our proposed HFL divides its connected remote devices into two categories: normal devices and stragglers, based on their communication and computation performance. This can be typically done by analyzing their historical behaviors. We denote the set of normal devices by $\mathcal{S}_1$ and the set of stragglers by $\mathcal{S}_2$, where $|\mathcal{S}_1| + |\mathcal{S}_2| = N$. Hence, the HFL consists of two components: synchronous kernel which communicates with the normal devices in each communication round, and the asynchronous updater which incorporates delayed model updates (i.e., neural network gradients) from stragglers.  

Similar to existing synchronous FL methods \cite{mcmahan2017communication, konevcny2016federated, li2019convergence}, we perform model update of the synchronous kernel as a weighted summation of a sequence of local updates.
For each straggler, we use $\tau_i$, $i \in \mathcal{S}_2$, to indicate the number of rounds behind to the current sequence. Especially, we have $\tau_i = 0$, $i \in \mathcal{S}_1$, for normal devices. 
Then, the current joint model with both the synchronous kernel and the asynchronous updater would be 
\begin{align}
\label{Eq:joint}
    \mathbf{w}_t = \left \{ \begin{array}{cc}
          \sum\limits_{i \in \mathcal{S}_1} p_i \mathbf{w}^{i}_{t} &  \text{if } t \leq \max\limits_{i \in \mathcal{S}_2} \tau_i \\ 
          \sum\limits_{i \in \mathcal{S}_1} p_i \mathbf{w}^{i}_{t} + \sum\limits_{i \in \mathcal{S}_2} p_i \mathbf{w}_{t-\tau_i}^{i} & \text{Otherwise}\\ 
    \end{array}
    \right .
\end{align}

Figure~\ref{Fig:structure} illustrates the joint learning process of the proposed HFL using both synchronous kernel and asynchronous updater with the presence of stragglers, where only one normal device and two stragglers are only given as an example. At $t=0$, the server initializes a joint model and broadcasts it to all devices. At $t=1$, only the normal device completes local training and return its model updates to the server for synchronously updating the joint model, handled by the synchronous kernel. At $t=2$ and $t=3$, when a straggler completes its local training, the asynchronous updater can incorporate the delayed model updates to the joint model.

It is clear that the main challenge is how to utilize the delayed model updates to contribute the training of the current joint model. The shown updating mechanism in Equation~(\ref{Eq:joint}) is problematic in that at the $t$-th communication round, the $i$-th straggler actually sends back a delayed gradient $g(\mathbf{w}_{t-\tau}^{i})$. To reach the optimal solution of $\mathbf{w}_{t}$, we need to approach the ``up-to-date" gradient $g(\mathbf{w}_{t}^{i})$ from $g(\mathbf{w}_{t-\tau}^{i})$, which is a well-known problem in asynchronous SGD optimization field \cite{stich2019error, arjevani2020tight, glasgow2020asynchronous}. In addition,  stragglers in a FL network can have a variety of computation power, which leads to a highly unbalanced distribution of $\tau$. Thus, in this paper, we develop an adaptive approximation solution to address this problem.

\subsection{Adaptive Delayed-SGD (AD-SGD)}\label{SubSec: DA-SGD}
In order to bridge the gap between the gradient from the joint model  $g (\mathbf{w}_{t}^{i})$ to the delayed gradient $g (\mathbf{w}_{t-\tau}^{i})$, we propose a novel adaptive approximation method. For simplicity, when there is no confusion, we omit the index $i$ in the rest of this paper for presentation purpose. 

\noindent\textbf{Taylor Expansion.} We apply the Taylor expansion \cite{folland2005higher, bischof1993structured} to expand the gradient update $g(\mathbf{w}_{t})$ for the current joint model at the $(t-\tau)$-th step as follows
\begin{equation}\label{Eq:Taylor}
\begin{split}
g (\mathbf{w}_{t}) =g (\mathbf{w}_{{t-\tau}}) +  \nabla g &(\mathbf{w}_{{t-\tau}})  (\mathbf{w}_{t} - \mathbf{w}_{t-\tau}) \\
& + \mathcal{O}((\mathbf{w}_{t} - \mathbf{w}_{t-\tau})^2)I_n,
\end{split}
\end{equation}
where $I_n$ is a $n$-dimension all-ones vector, and $\nabla g (\cdot) $  represents a gradient matrix, whose element $g_{i, j} = \frac{\partial \mathcal{L}^2}{\partial w_i \partial w_j}$ for $i, j \in n$.  Note that $g(\mathbf{w}_{t-\tau})$ is the zero-order item in the Taylor expansion of $ g(\mathbf{w}_{t})$ and the major difference between the expected model gradient $g(\mathbf{w}_{t})$ and the delayed model gradient $g (\mathbf{w}_{t-\tau})$ comes from the higher-order components $ \nabla g(\mathbf{w}_{t-\tau}) (\mathbf{w}_{t} - \mathbf{w}_{t-\tau}) + \mathcal{O}((\mathbf{w}_{t} - \mathbf{w}_{t-\tau})^2)I_n$. Intuitively, we could use the full Taylor expansion in Equation~\eqref{Eq:Taylor} to approach $g(\mathbf{w}_{t})$, however, this is unrealistic due to its high computation cost. Additionally, even solving the first-order item $ \nabla g(\mathbf{w}_{t-\tau}) (\mathbf{w}_{t} - \mathbf{w}_{t-\tau})$ is also highly non-trivial, which is considered to be the Hessian matrix $H(\mathbf{w}_{t-\tau})$ of the joint loss function. 

\noindent\textbf{Approximation of Hessian Matrix.} Since the computation cost of the Hessian matrix in the FL network is still expensive, we consider an alternative approach to address this problem with a limited computational resource.

In particular, we use the gradient of the joint model $g(\mathbf{w}_{t-\tau})$ computed during the local training process. An outer product matrix $R(\mathbf{w}_{t - \tau}) \in \mathbb{R}^{n \times n}$ can be obtained with  $g(\mathbf{w}_{t-\tau})$ at the $(t-\tau)$-th communication round as 
\begin{equation}\label{Eq:Outer_matrix}
R(\mathbf{w}_{t-\tau}) = \left( \frac{\partial}{\partial \mathbf{w}} \mathcal{L}(\mathbf{w}_{t-\tau}; \mathbf{X}) \right) \left( \frac{\partial}{\partial \mathbf{w}} \mathcal{L}(\mathbf{w}_{t-\tau}; \mathbf{X}) \right)^\top. \end{equation}

It can be seen that the outer product matrix $R(\mathbf{w}_{t - \tau})$ and the Hessian matrix ${H}(\mathbf{w}_{t - \tau})$ are two equivalent methods to calculate the fisher information matrix \cite{friedman2001elements}, because the cross entropy loss in this case is a negative log-likelihood with respect to the softmax function.
This equivalent approach for solving the Hessian matrix has been applied in the recent  works \cite{choromanska2015loss, kawaguchi2016deep}. Thus, from the already computed gradients, we obtain an alternate approximation method to the Hessian matrix by the outer production. 

\noindent\textbf{Adaptive Hyper-parameter.} Using the gradient outer product to approximate the Hessian matrix, the optimal gradient $g(\mathbf{w}_{t})$ could be represented as follows
\begin{equation}\label{Eq:newgrad}
    g (\mathbf{w}_{t}) = g (\mathbf{w}_{t-\tau}) + R(\mathbf{w}_{t-\tau}) (\mathbf{w}_{t} - \mathbf{w}_{t-\tau}).
\end{equation}
However, this approximation still could still a large error from the omitted high-order items $\mathcal{O}((\mathbf{w}_{t}) - \mathbf{w}_{t-\tau})^2)I_n$ in the Taylor expansion, especially with a large delay $\tau$, which is common in FL network settings. Thus, in order to reduce the impact from different stragglers, we introduce an adaptive hyper-parameter to control the weight of our approximation into the joint model aggregation, which is related to the value of $\tau$ and the training round $t$. The main idea is that a slower straggler (i.e., a larger $\tau$) contributes less into the joint model training, and this contribution continuously decreases when communication round increasing (i.e., $t$) in order to reduce oscillation when the learning converges. In particular, our AD-SGD method introduces a hyper-parameter $\lambda_{t}$ with an exponential decay function on $t$ and $\tau$ as follows
\begin{equation}\label{Eq:Lambda}
\lambda_t =\lambda_{0} \exp(-(t-\tau)). 
\end{equation}
where $\lambda_{0}$ is a user-defined parameter. While its optimal value can be determined through cross validation, a constant value of $\lambda_{0} = 0.5$ is used in our experiments. The choice of $\lambda_0$ in our HFL algorithm will also be discussed in detail with experimental results in Section.~\ref{SubSec:convex}. 

Therefore, using the AD-SGD method, the joint model update for our HFL algorithm in Equation~\eqref{Eq:joint} can be given
\begin{equation}\label{Eq:New}
\mathbf{w}_{t} = (1- \lambda_t) \hat{\mathbf{w}}_t + \lambda_t \sum_{i \in \mathcal{S}_2 } p_i (\mathbf{w}_{t-\tau_i}^{i} - \eta_t  g (\mathbf{w}_{t}^{i})),
\end{equation}
where $g(\mathbf{w}_{t}^{i})$ is approximated from the outer product matrix as $g (\mathbf{w}_{t}^{i}) =g (\mathbf{w}_{t-\tau_i}^{i}) + R(\mathbf{w}_{t-\tau_i}^{i}) ({\mathbf{w}}_{t}^{i} - \mathbf{w}_{t-\tau_i}^{i}).$  

\subsection{Algorithm Description}
To this end, we use the AD-SGD method to bridge the gap between the delayed gradients and the optimal gradients for  stragglers in FL. In particular, we approximate the Hessian matrix using the outer product gradient with a low computation cost, which have been proved to be equivalent for the calculation of the Fisher information matrix. With the help of AD-SGD, we are able to achieve a learning balance in efficiency and effectiveness when stragglers are present in FL. Specifically, we summarize the proposed HFL algorithm in Algorithm~\ref{alg:device} and \ref{alg:server}, where Algorithm~\ref{alg:device} introduces the local learning process and Algorithm~\ref{alg:server} shows the joint model updating mechanism.  

According to Algorithm~\ref{alg:device}, at the $t$-th round, the $i$-th remote device receives the current joint model $\mathbf{w}_{t}$ from the server, performs the local training process with $E$ epochs and sends model updates back to the server at the $(t + \tau_i)$-th round with various delays for stragglers ($\tau_i \geq 0$). And for the server side, unlike existing FL methods, the server stores a backup model at the $t$-th round when the straggler receives the joint model. When the delayed gradient $g(\mathbf{w}_{t+ \tau_i}^{i})$ is received by the server at the $(t+\tau)$-th round, the server updates the joint model based on the updating rule using Equation~\eqref{Eq:New}.

Note that compared to existing synchronous FL algorithms, e.g., FedAvg, there is no extra communication rounds and extra remote device computational requirement in our HFL. In particular, the approximation cost from the Equation~\eqref{Eq:New} mainly comes from the additional storage of several previous joint models on the server. The backup of the previous joint model does not violate the privacy settings of the FL network. 

\begin{algorithm}[tb]
	\caption{HFL: Remote device side, index $i$}
	\label{alg:device}
	\begin{algorithmic}
		\STATE {\bfseries Input:} Training data $\mathbf{X}^i$, local epoch number $E$.
		\STATE The $i$-th device receives the learning rate $\eta_t$ and the latest joint model $\mathbf{w}_{t}$ from the server at the $t$-th round.
		\STATE Initialize local model: $\mathbf{w}_t^i = \mathbf{w}_t$.
		\FOR{$e=0$ {\bfseries to} $E-1$}
		\STATE Perform local training process via SGD using Equation~\eqref{Eq:local}.
		\ENDFOR
		\IF {Communication round is $(t+\tau_i)$}
		\STATE Return $\mathbf{w}_{t+\tau_i}^{i}$ and $g (\mathbf{w}_{t+\tau_i}^{i})$ to the server.
		\ENDIF
	\end{algorithmic}
\end{algorithm}

\begin{algorithm}[tb]
	\caption{HFL: Server side}
	\label{alg:server}
	\begin{algorithmic}
		\STATE {\bfseries Initialization:}  FL network with $N$ devices, communication round $T$, initialized parameter $\lambda_0$, and two sets of remote devices $\mathcal{S}_1$ for stragglers and $\mathcal{S}_2$ for normal devices.
		\FOR{$t=0$ {\bfseries to} $T-1$}
		    \IF{Updates received from $i$-th device}
		        \IF{$i \in \mathcal{S}_1$}
		            \STATE Update $\mathbf{w}_{t}$ with local model updates $\mathbf{w}_{t-\tau_i}^{i}$ and $g (\mathbf{w}_{t-\tau_i}^{i})$.
	            \ELSIF{$i \in \mathcal{S}_2$}
	                \STATE Recall the saved joint model $\mathbf{w}_{t-\tau_i}$.
	                \STATE Approximate the optimal update $g (\mathbf{w}_{t}^{i})$ using Equation~\eqref{Eq:newgrad}.
	                \STATE Update the $\mathbf{w}_{t}$ using  Equation~\eqref{Eq:New}.
		        \ENDIF
		        \STATE Broadcast the updated $\mathbf{w}_{t}$ and $\eta_t$ to the $i$-th device.
		    \ENDIF
		\ENDFOR
	\end{algorithmic}
\end{algorithm}

\section{Convergence Analysis} \label{Sec:Convergence}
In this section, we provide the convergence analysis of the proposed HFL algorithm with the adaptive approximation method AD-SGD for the delayed gradients. In this paper, both the convex and non-convex optimization problems are investigated. Due to the space limitation, we only provide the results and leave the detailed proof to the supplementary materials in Section~B.

\subsection{Assumptions}
To illustrate the convergence analysis, we first provide several assumptions, which are widely used in the previous works on FL \cite{stich2018local, sahu2018federated, li2019convergence, khaled2020tighter, karimireddy2020scaffold}.

\begin{assumption}\label{Assum:Global_2}
	\textbf{\emph{(L-smooth):}} The learning objective $F(\cdot)$ is $L$-smooth with $L \geq 0$ such that 
	\begin{equation}\label{Eq:L_smooth}
	||\nabla F(\mathbf{v}) -\nabla F(\mathbf{u})|| \leq L || \mathbf{v}- \mathbf{u} ||, ~\forall \mathbf{u}, \mathbf{v}.
	\end{equation}
\end{assumption}

\begin{remark}
    When the objective is $\mu$-convex and satisfies $L \geq \mu$, the results in Assumption~\ref{Assum:Global_2} leads to 
	\begin{equation}
	\frac{1}{2L}||\nabla F (\mathbf{v})||^2_2 + \frac{1}{\mu} ||\mathbf{v} - \mathbf{u}||^2_2	 \leq \langle \mathbf{v}- \mathbf{u}, \nabla F(\mathbf{u})\rangle. 		
	\end{equation}
\end{remark}

\begin{assumption}\label{Assum:Global_4}
	\textbf{\emph{(Bounded gradient):}} We assume that the delayed gradients in the HFL algorithm are uniformly bounded as 
	\begin{equation}\label{Eq:Uni_bound2}
	\mathbb{E} ||g (\mathbf{w}_{t})||^2 \leq G^2 
	\end{equation}
\end{assumption}

\begin{definition}\label{Definition_1}
	\textbf{\emph{(Local dissimilarity):}} We define the difference between the $i$-th local objective $F_i(\cdot)$ and the joint objective $F(\cdot)$ with the same joint model $\mathbf{w}_{t}$ is bounded as 
\begin{equation}
	\mathbb{E} ||\nabla F_i (\mathbf{w}_{t})||^2 \leq B^2 ||\nabla F (\mathbf{w}_{t})||^2.
	\end{equation}
\end{definition}
Note that when $B=1$, there is a special case that the local objective is the same as the joint objective. In this paper, we consider the scenario that $B > 1$. Additionally, to quantify the heterogeneity of the FL network, we also introduce the non-i.i.d. degree of the learning data respect to the weighted factor $p_i$. For the normal devices in the synchronous kernel, we represent the learning data distribution as $\Psi_1 $, where $\Psi_1 = \sum_{i \in \mathcal{S}_1} p_i.$ Similarly, the distribution for the stragglers is denoted as $\Psi_2 = \sum_{i \in \mathcal{S}_2} p_i.$ Obviously, we have $\Psi_1 + \Psi_2 = \sum_{i=1}^{N} p_i = 1.$ 

\subsection{Optimization Analysis}

\begin{theorem}\label{Theorem:1}
\textbf{\emph{(Convex HFL convergence):}}  
    For the convex optimization problems, let the Assumptions in this paper hold that $F(\cdot)$ is $\mu$-convex and L-smooth,  our HFL algorithm satisfies \footnote{The detailed proof is shown in the supplementary material Section $B.1$.} 
\begin{equation}
\mathbb{E} F(\mathbf{w}_{t}) -  F(\mathbf{w}^{\star}) \leq   \frac{ L_2 L^3 \tau^2 G^2 \sigma^2 }{\mu^6 (t+\tau)^2 B_2} +  \frac{L^3 G^2}{2 (t+\tau) \mu^4 B_2}, \nonumber
\end{equation}
where $B_2 = B^8 \Psi_1^2 \Psi_2^2$, and $\eta_{t}$ is chosen to satisfy $\eta_t \leq \frac{L}{\mu^2 t B^4 \Psi_1 \Psi_2}$.
\end{theorem}

\begin{corollary}
\textbf{\emph{(Convergence rate convex):}} When the convergence of the convex problem is guaranteed by the settings $\eta_t \leq \frac{L}{\mu^2 t B^4 \Psi_1 \Psi_2},$ the optimization bound comes from two parts: a high-order part $\frac{ L_2 L^3 \tau^2 G^2 \sigma^2 }{\mu^6 (t+\tau)^2 B_2}$ and a low-order part $\frac{L^3G^2}{2(t+\tau) \mu^4 B_2}$. Note that the high-order term would converge to a stationary point faster as $t$ grows and we consider the convergence rate of HFL against convex problems as follows
\begin{equation}
    \mathbb{E} F(\mathbf{w}_{t}) -  F(\mathbf{w}^{\star}) \leq  \mathcal{O}(\frac{1}{t+\tau}).
\end{equation}

\end{corollary}

\begin{theorem}\label{Theorem__1}
    For the non-convex problems under the Assumption 1-2, we consider the model convergence with a constant learning rate $\eta_t$ that  \footnote{The detailed proof is shown in the supplementary material Section $B.2$. }
\begin{equation}
	\begin{split}
	\min_{t \in T} \mathbb{E}& || \nabla F(\mathbf{w}_t)|| ^2 \\
	& \leq  \frac{1}{T\eta_t B_1} \left[ F(\mathbf{w}_{0}) -  F(\mathbf{w}^{\star}) ) \right],
	\end{split}
\end{equation}
\end{theorem}
where the maximum of $\tau$ is bounded as $t+ \tau \leq T$, $B_1 = B^4 \Psi_1 \Psi_2$ and the constant value of $\eta_t$ satisfies the following inequality that 
\begin{equation}\label{Eq:condition2}
    \frac{\eta_{t}^2 LG^2 \sqrt{\frac{\Psi_2 B_1}{\Psi_1}}}{2} - \eta_{t}B_1G^2 \leq 0.
\end{equation}

\begin{corollary}
\textbf{\emph{(Convergence rate non-convex):}} Let $\eta_t \leq \frac{2}{L} \sqrt{\frac{\Psi_1 B_1}{\Psi_2}},$ then the inequality in Equation~\eqref{Eq:condition2} is satisfied and we have the convergence rate of our HFL algorithm against non-convex problems as follows
\begin{equation}\label{Eq:rate_non-convex}
    \min_{t \in T} \mathbb{E} || \nabla F(\mathbf{w}_t)|| ^2  \leq \mathcal{O}(\frac{1}{T}).
\end{equation}
\end{corollary}

\textbf{Discussion.}
Following the above proof steps, we provide the convergence guarantee and obtain the convergence rate for our HFL algorithm. Let the maximum delayed gradient $\tau$ is bounded by $t + \tau \triangleq T$, the convergence rate for both convex and non-convex optimization problems is $\mathcal{O}(\frac{1}{T}).$ Recall that in the FedAvg for non-i.i.d. data learning problems \cite{li2019convergence}, the convergence rate is $\mathcal{O}(\frac{1}{t})$, where $t$ is the synchronous communication round. We would note that the convergence rate of the SOTA FedAvg algorithm could be regarded as a special case in our HFL when $\tau = 0$, which also follows the design motivation of our algorithm. 

Moreover, the magnitude of the communication round for a straggler is $\frac{T}{\tau}$, which indicates the communication cost in the HFL algorithm is still close to $T$. Specifically, when $\tau$ gets larger, the convergence rate of HFL might be slower. Which indicates that with fixed  $T$ communication rounds. the learning performance of HFL can be decreased as the value of $\tau$ grows. And we will show empirical results for the choice of different $\tau$ in Section~\ref{Sec:Experiments}.

\section{Experiments}\label{Sec:Experiments}

\subsection{Experimental Setup}

\begin{figure*}[t!]
	\centering
	\begin{subfigure}{0.32\columnwidth}
		\includegraphics[width = 1\columnwidth]{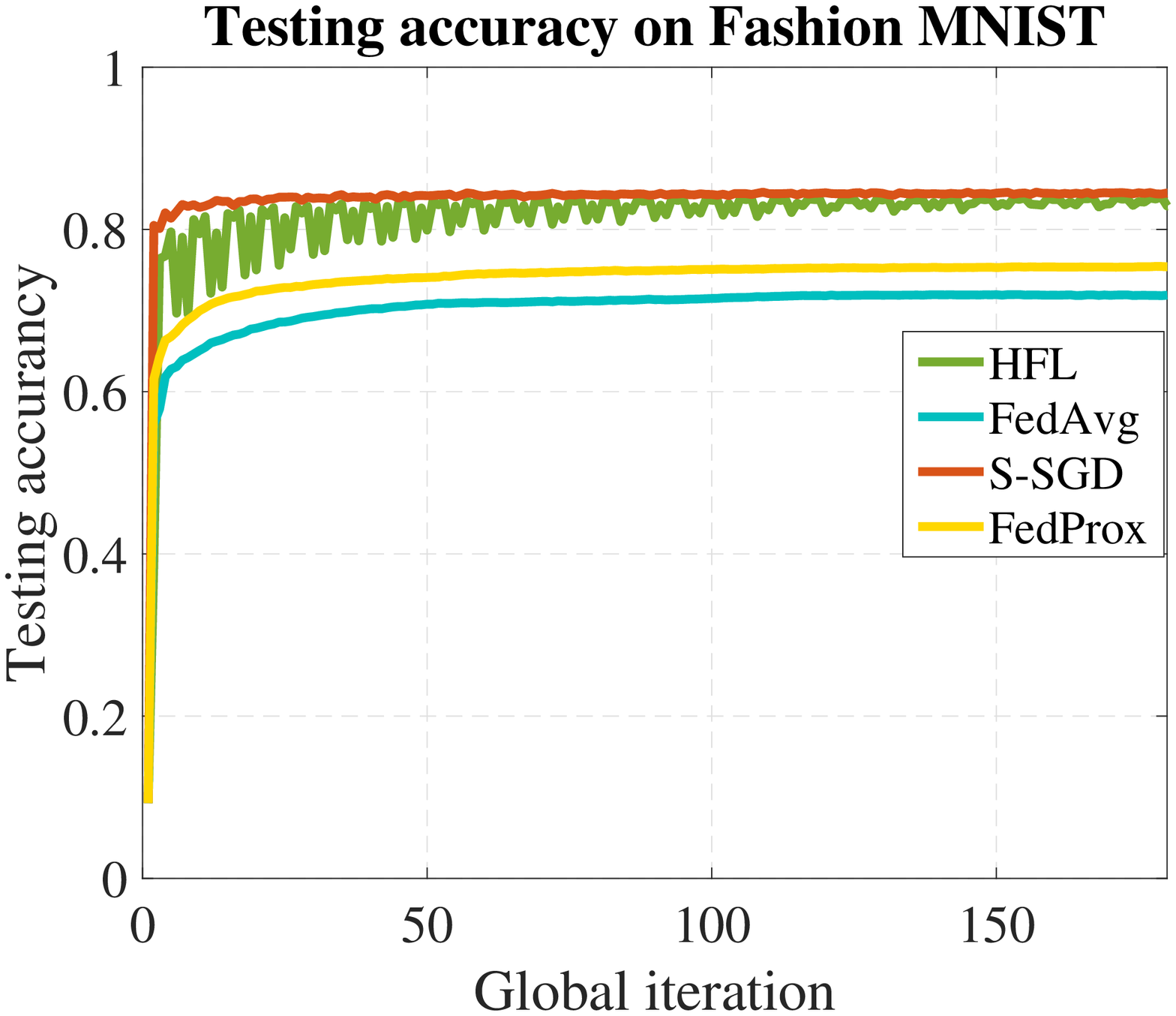}
 		\caption{}
		\label{fig:fmnistnoniid_acc}
	\end{subfigure}
	\begin{subfigure}{0.32\columnwidth}
		\includegraphics[width = 1\columnwidth]{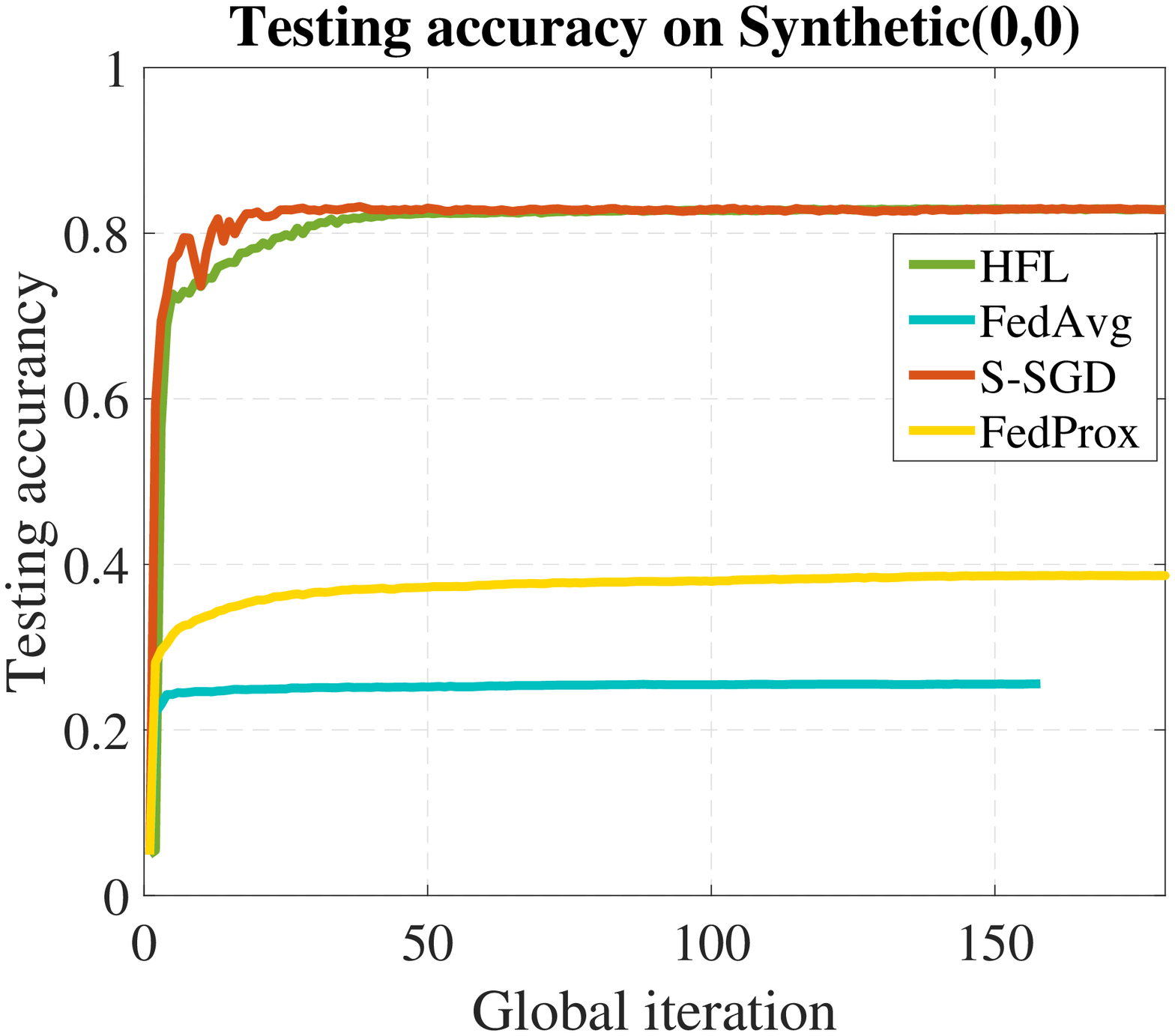}
 		\caption{}
		\label{fig:synthetic00noniid_acc}
	\end{subfigure}
	\begin{subfigure}{0.32\columnwidth}
		\includegraphics[width = 1\columnwidth]{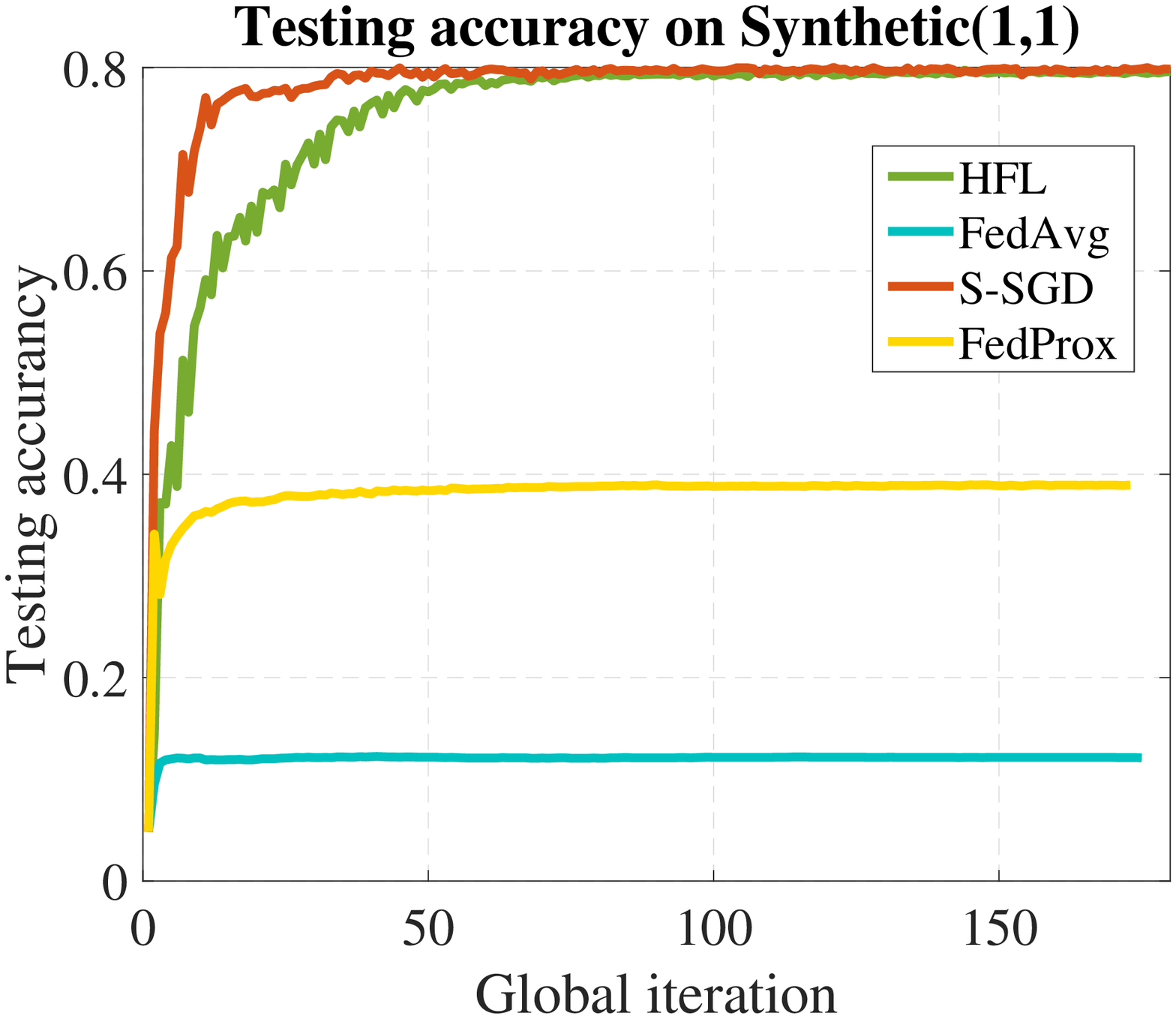}
 		\caption{}
		\label{fig:synthetic11noniid_acc}
	\end{subfigure}
	\caption{The learning accuracy comparison of our HFL to the excising methods on the convex datasets: a) Fashion MNIST; b) $Synthetic(0,0)$; c) $Synthetic(1,1)$.}
	\label{fig:noniid}
\end{figure*}

\noindent\textbf{Models and Datasets.}
To evaluate the performance of the proposed HFL algorithm, we conduct experiments on multiple datasets for both convex and non-convex optimization problems. For convex optimization, we design experiments with logistic regression models on the Fashion MNIST \cite{xiao2017/online} and a synthetic dataset. We distribute training samples to different remote devices following the power law distribution to obtain non-i.i.d distributed learning data. 
The synthetic dataset is generated in the same way as presented in previous studies \cite{li2019convergence}, which is easy to be manipulated for controlling data heterogeneity.   
We denote it as $Synthetic(\gamma, \xi)$, where $\gamma$ controls the difference between any two local models and $\xi$ controls how much difference between the learning data in each device. For non-convex optimization, we select Sentiment140 (Sent140) \cite{go2009twitter} and The Complete Works of William Shakespeare  (Shakespeare) \cite{mcmahan2017communication} with a LSTM classifier.   

\noindent\textbf{Experimental Settings.}
All experiments are conducted in a distributed network that consists of $N = 100$ devices. The total number of global communication is $T = 200$ and the number of local epochs is $E= 5$. The maximum value of $\tau$ is set to $10$ and $\tau_i \in \mathcal{S}_2$ are uniformly distributed. The following methods are used for comparison:

\textbf{1) Sequential SGD (S-SGD) \cite{zinkevich2010parallelized}.} We implement the standard S-SGD in a centralized environment. Therefore, this method provides the upper bound performance over all compared methods in this paper.

\textbf{2) FedAvg.} FedAvg is considered as one of the groundbreaking works in FL research field. We setup the FedAvg algorithm based on the settings in \cite{li2019convergence}, which provides a convergence guarantee against non-i.i.d. data problems. In particular, we set the possibility for the server to the remote devices corresponding to a normalized vector that $p_i$ is linear to $\tau_i$. The value of $K$ for the number of selected devices in each communication round is set to $10$ by default.

\textbf{3) FedProx \cite{sahu2018federated}.} FedProx is one popular variant of FedAvg which adds a quadratic proximal term to limit the impact from local updates in a heterogeneous network. In this paper, we follow the instructions provided in the original paper to evaluate the performance with $90\%$ stragglers and the same remote devices possibility distribution in FedAvg. 

\textbf{Implementations.} we define the joint model initialization $\mathbf{w}_{0} = 0$ and the initial learning rate $\eta_0 = 0.1$ with a decay function $\eta_{t} = \frac{\eta_0}{1+t}$. In each local training step, we consider that the remote device use SGD to train local model with its all local training data and with a batch size of $64$. For the delayed gradients in Equation~\eqref{Eq:Lambda}, we set $\lambda_0 = 0.5$.

\begin{figure}[t!]
	\centering
	\begin{subfigure}{0.49\columnwidth}
		\includegraphics[width = 1\columnwidth]{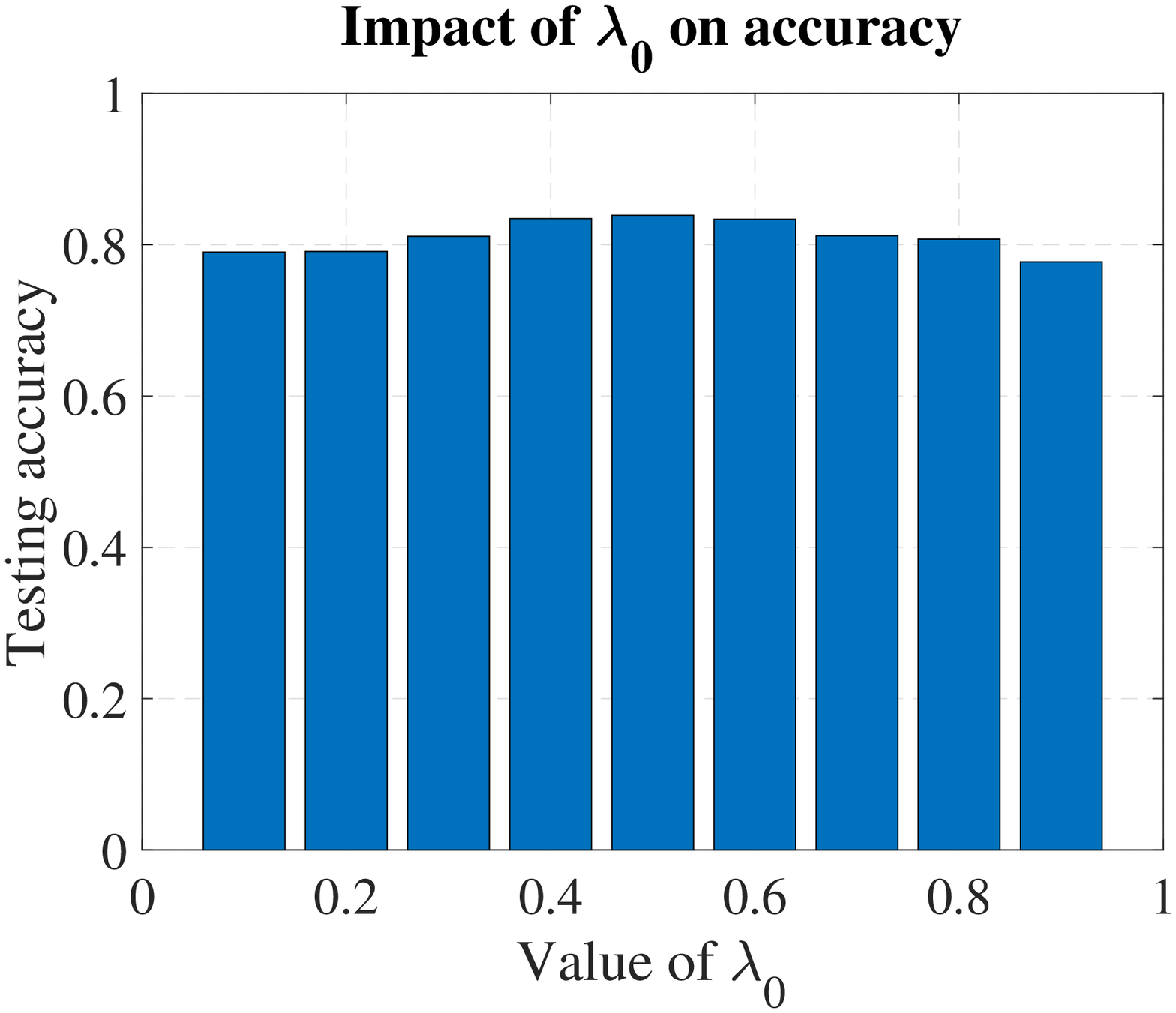}
		\caption{}
		\label{fig:lambda_acc}
	\end{subfigure}
	\begin{subfigure}{0.49\columnwidth}
		\includegraphics[width = 1\columnwidth]{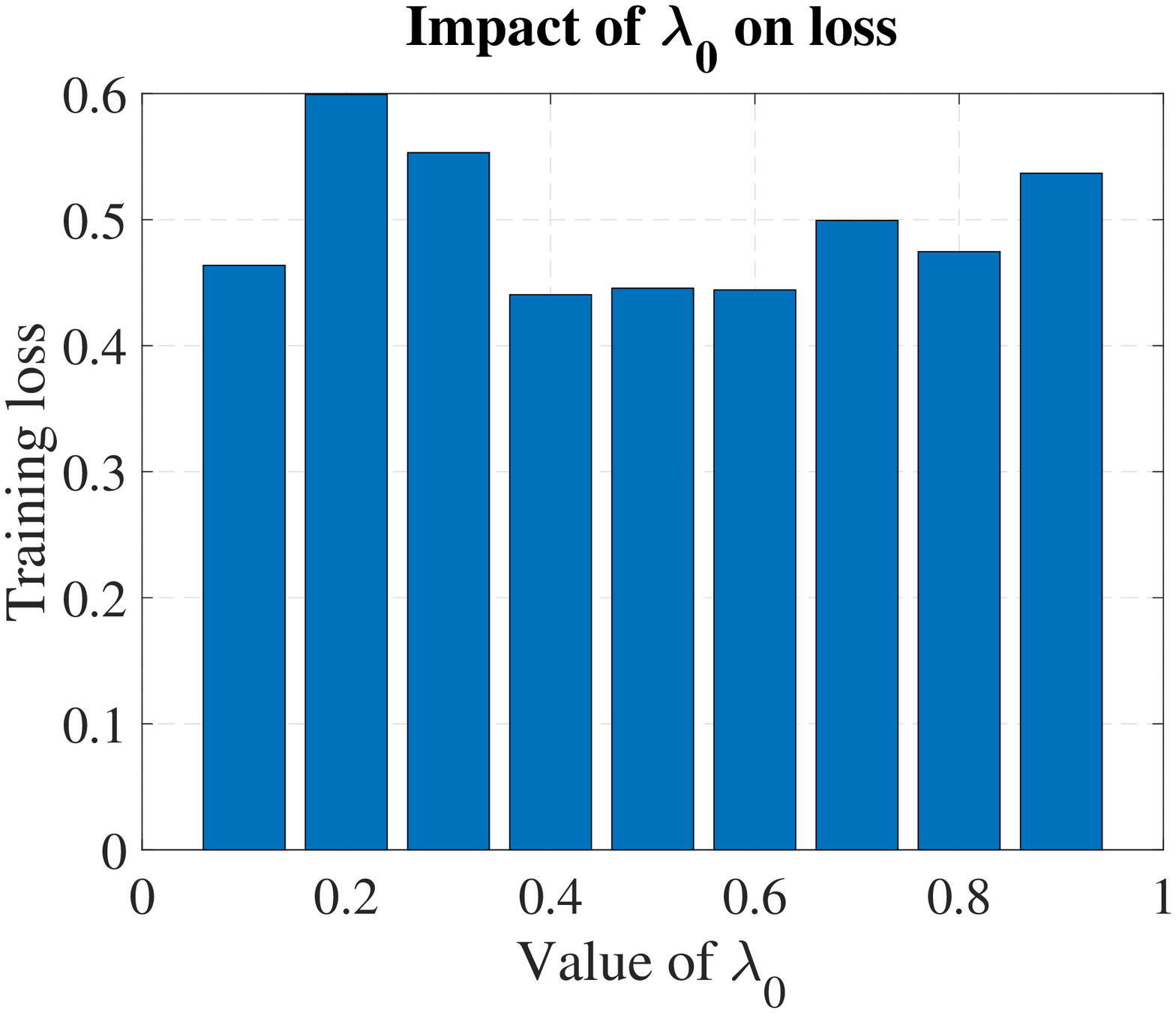}
		\caption{}
		\label{fig:lambda_loss}
	\end{subfigure}
	\caption{Choice of $\lambda_0$: a) testing accuracy of HFL with different $\lambda_0$ values; b) training loss of HFL with different $\lambda_0$ values.}
	\label{fig:lambda}
\end{figure}

\begin{figure}[t!]
	\centering
	\begin{subfigure}{0.49\columnwidth}
		\includegraphics[width = 1\columnwidth]{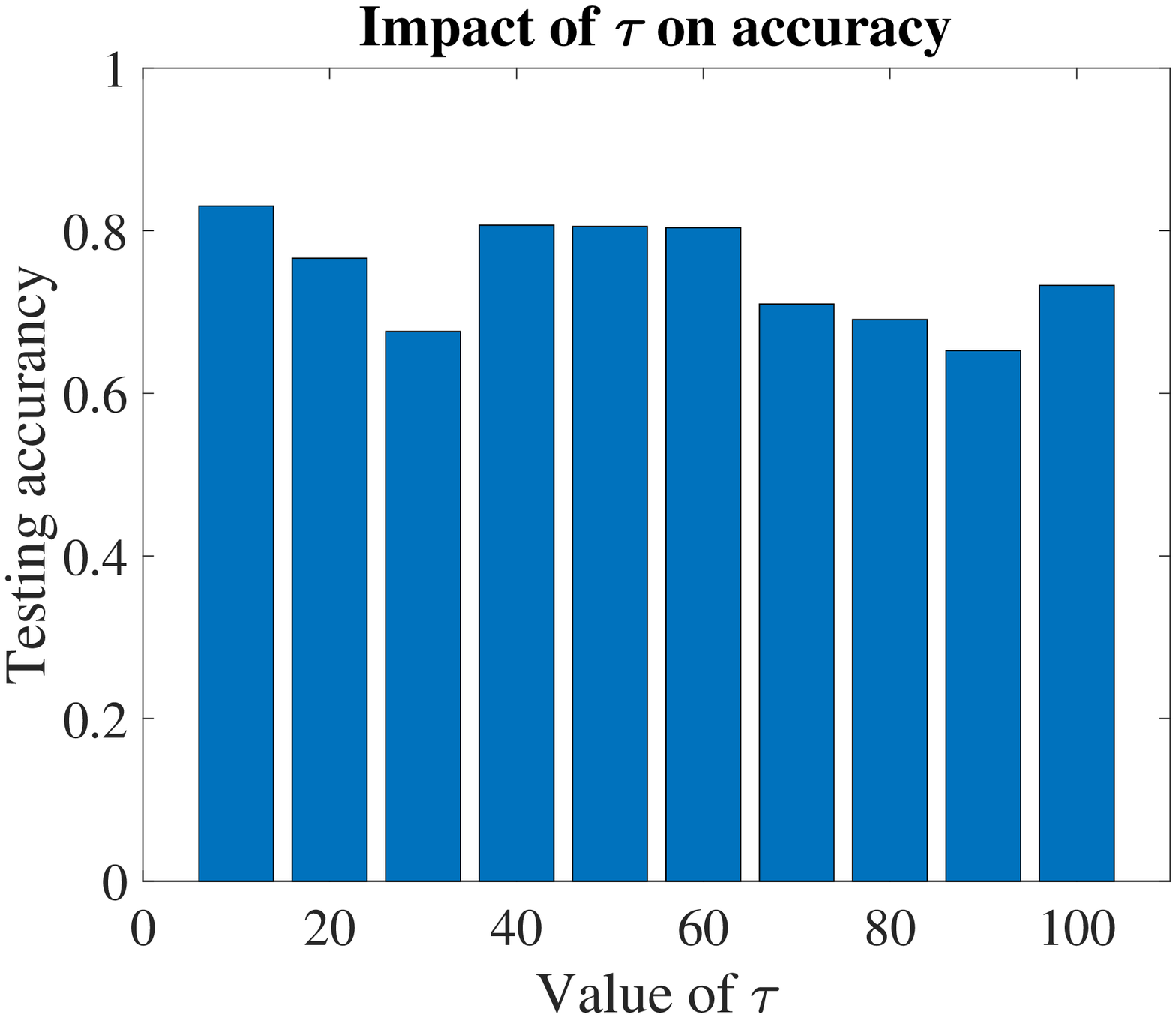}
		\caption{}
		\label{fig:tau_acc}
	\end{subfigure}
	\begin{subfigure}{0.49\columnwidth}
		\includegraphics[width = 1\columnwidth]{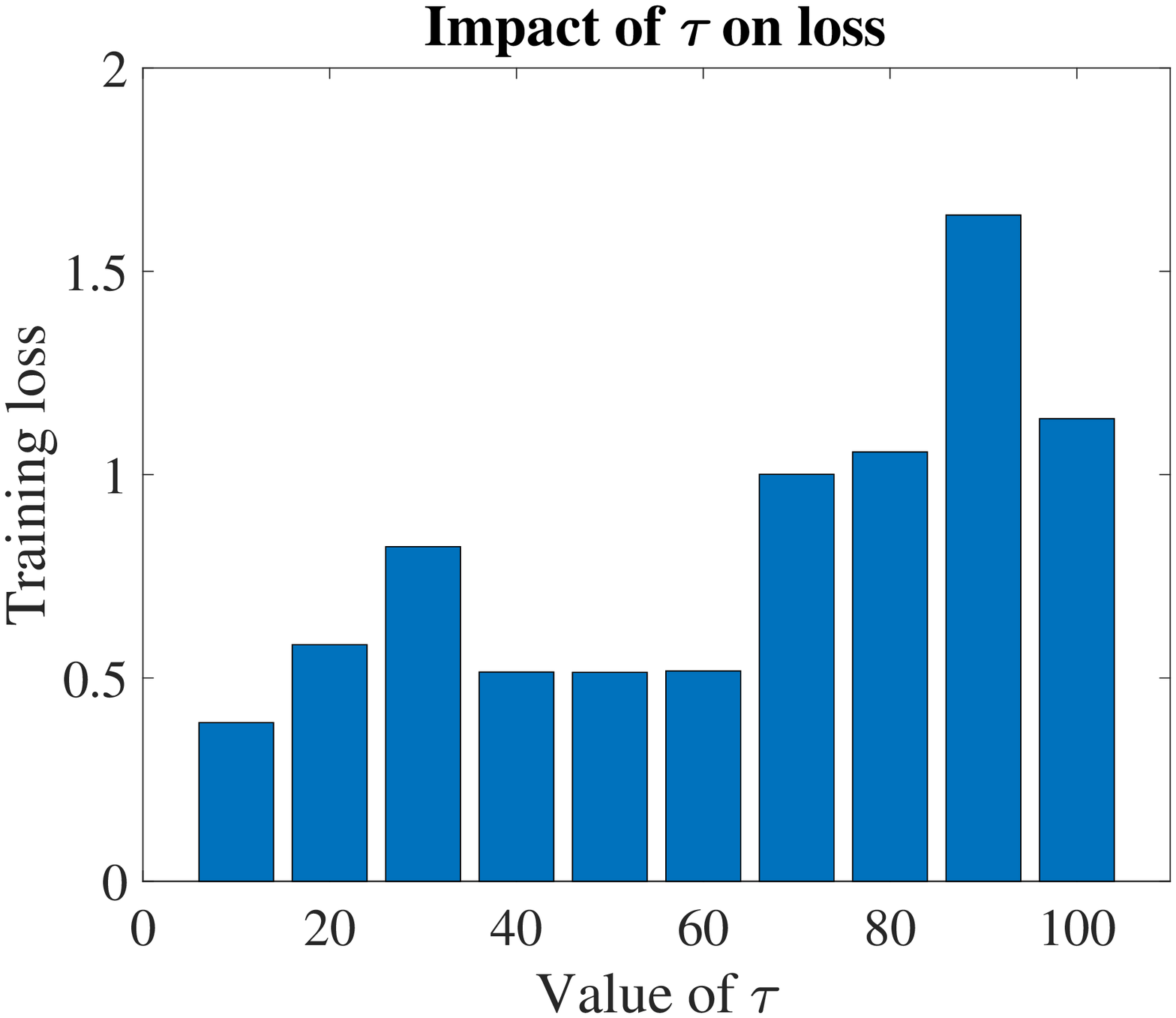}
		\caption{}
		\label{fig:tau_loss}
	\end{subfigure}
	\caption{Choice of maximum $\tau$: a) testing accuracy of HFL with different maximum $\tau$ values; b) training loss of HFL with different maximum $\tau$ values.}
	\label{fig:tau}
\end{figure}

\begin{figure*}[t!]
	\centering
	\begin{subfigure}{0.49\columnwidth}
		\includegraphics[width = 1\columnwidth]{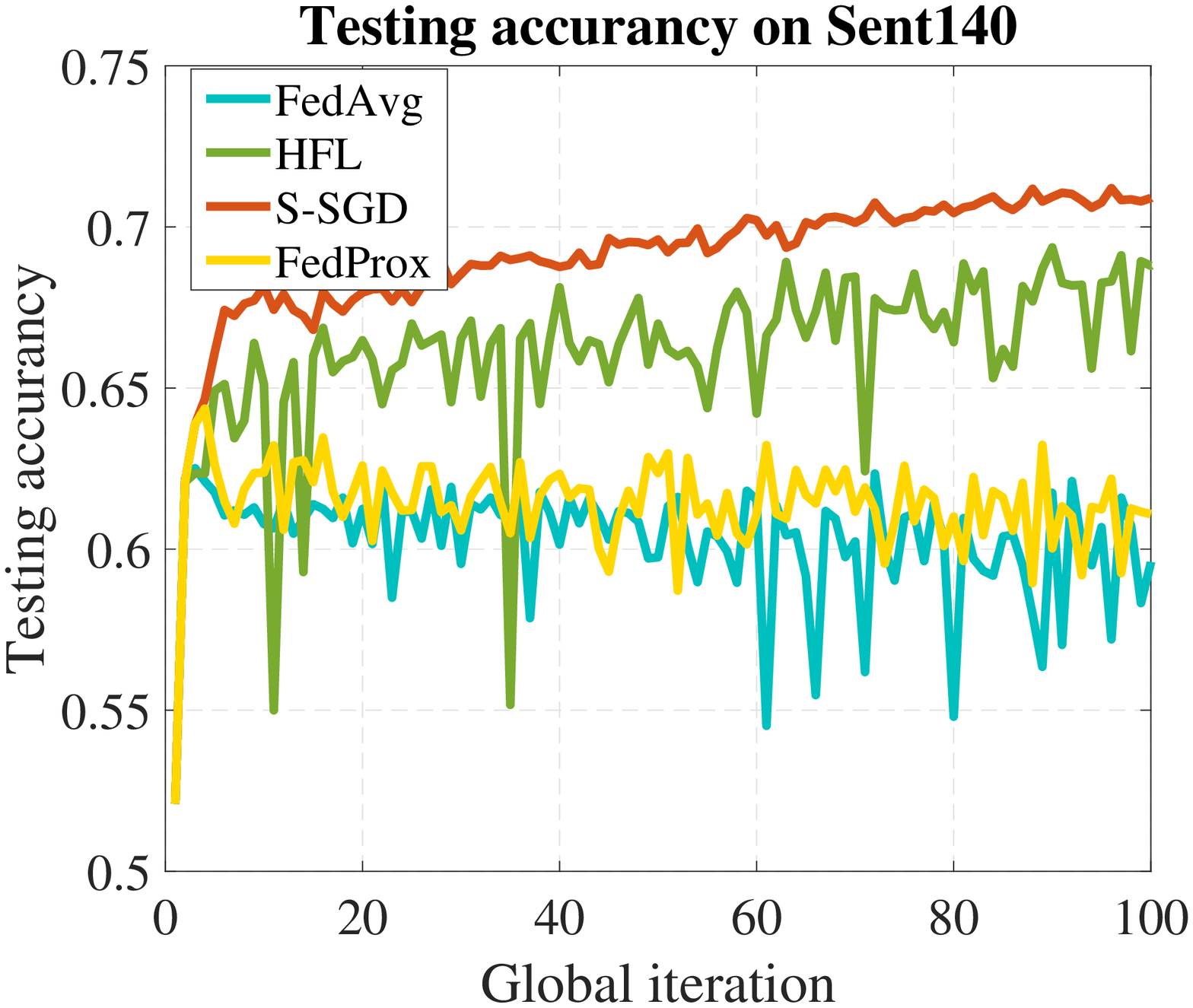}
		\caption{}
		\label{fig:sent140acc}
	\end{subfigure}
	\begin{subfigure}{0.49\columnwidth}
		\includegraphics[width = 1\columnwidth]{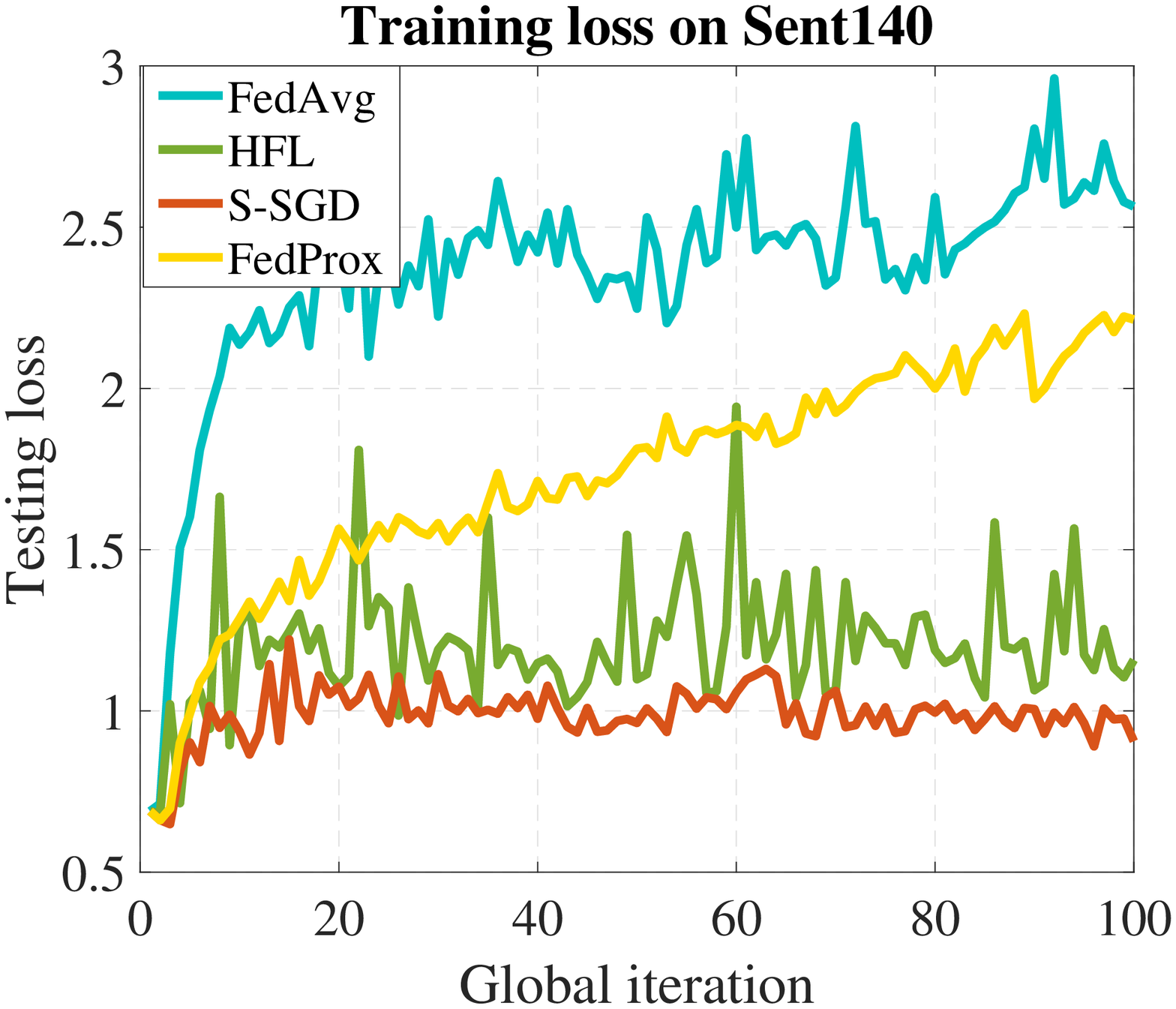}
		\caption{}
		\label{fig:sent140loss}
	\end{subfigure}
	\begin{subfigure}{0.49\columnwidth}
		\includegraphics[width = 1\columnwidth]{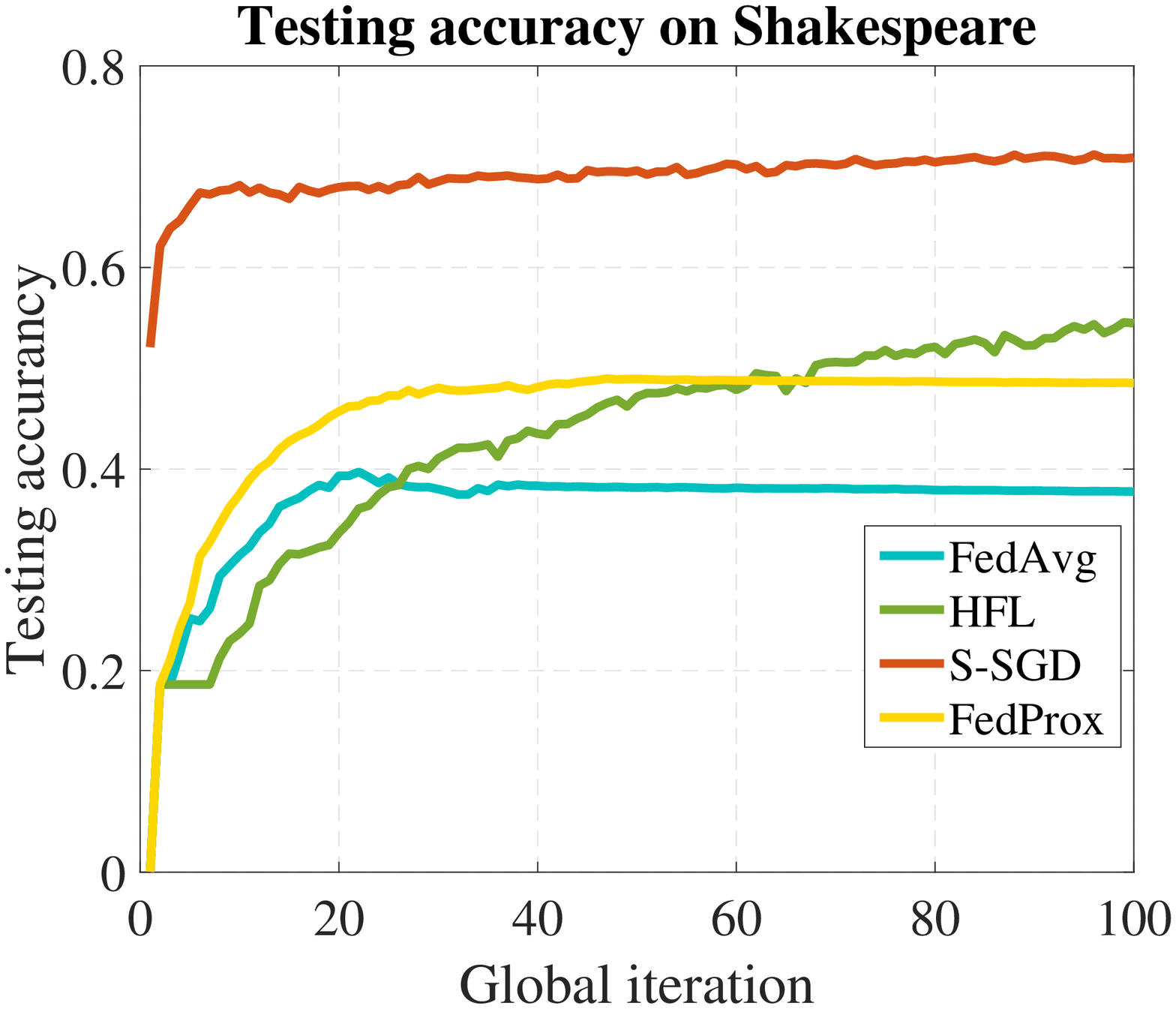}
		\caption{}
		\label{fig:shakeacc}
	\end{subfigure}
	\begin{subfigure}{0.49\columnwidth}
		\includegraphics[width = 1\columnwidth]{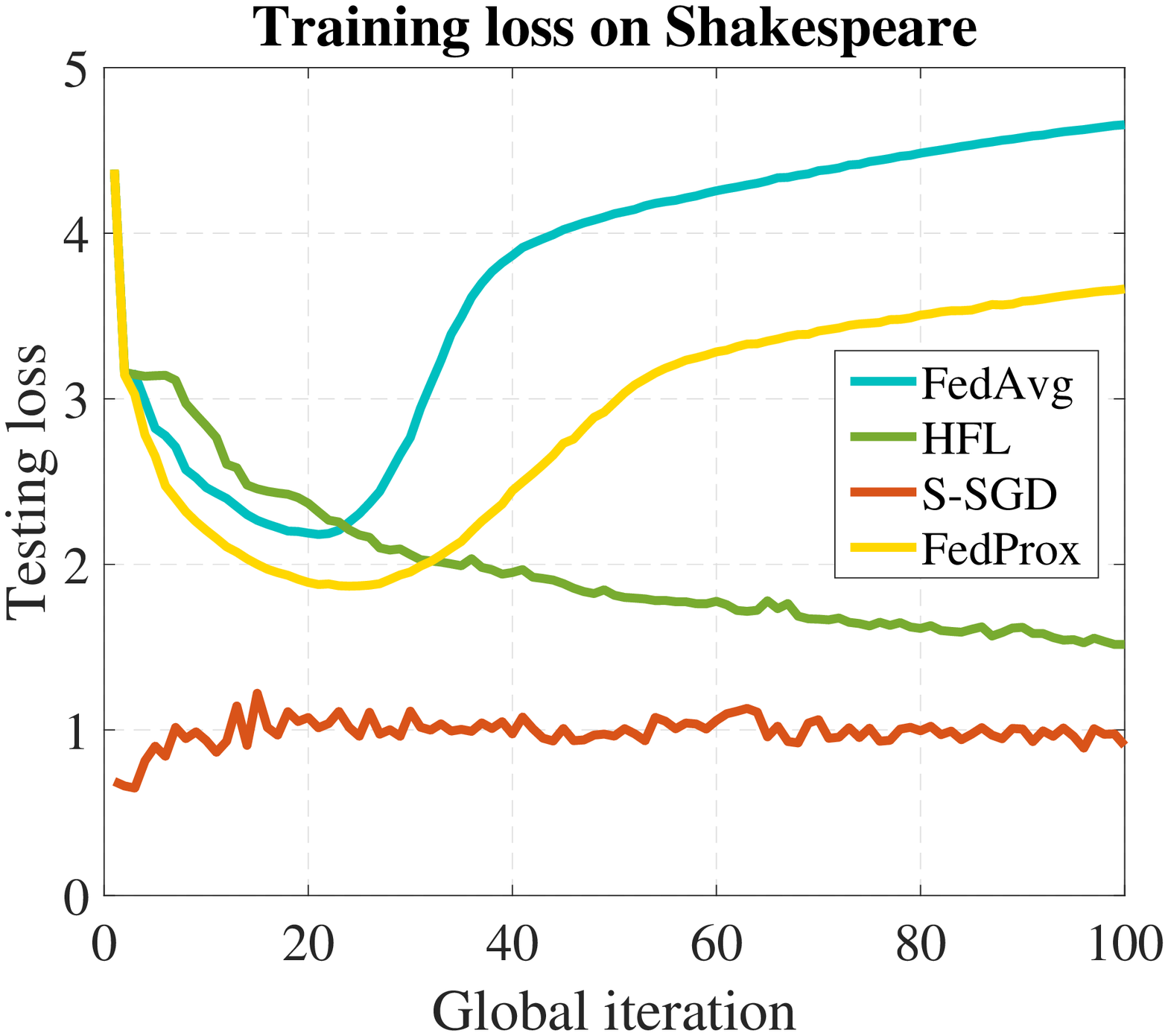}
		\caption{}
		\label{fig:shakeloss}
	\end{subfigure}
	\caption{The learning performance of the proposed HFL compared with existing methods on the non-convex datasets: a) testing accuracy of Sent140; b) training loss of Sent140; c) testing accuracy of Shakespeare; d) training loss of Shakespeare.}
	\label{fig:nonconvex}
\end{figure*}
\subsection{Convex Optimization Results}\label{SubSec:convex}

\textbf{Comparison to benchmarks.}
In Figure.~\ref{fig:noniid}, we compare learning performance for the convex optimization on both synthetic and Fashion MNIST datasets with logistic regression. The results show that our HFL algorithm achieves the best overall testing accuracy compared to other methods. For the Fashion MNIST dataset, it can be seen that the upper bound testing accuracy for Fashion MNIST using the S-SGD method is $84.58\%$ and our HFL reaches $83.41\%$ testing accuracy. Meanwhile, the converged accuracy for FedAvg is only $71.92\%$ and $75.49\%$ for FedProx. Note that the learning curve of HFL contains some obvious oscillations, which mainly comes from the approximation of the optimal model from delayed gradients. We can also notice that as the learning steps $t$ grows, the amplitude of the oscillation becomes smaller, which is mainly due to the use of our AD-SGD method that can reduce the impact of delayed gradients in an adaptive way. 

For Synthetic dataset, we evaluate the performance of HFL under two $\gamma$ and $\xi$ settings, i.e., $Synthetic(0,0)$ and $Synthetic(1,1)$. The results are shown in Figures~\ref{fig:synthetic00noniid_acc}-\ref{fig:synthetic11noniid_acc}. We could notice that compared to the results in the Fashion MNIST dataset, the performance improvement of HFL is greater than other methods on both two synthetic datasets. Meanwhile, the convergence rate of HFL is slower in $Synthetic(1,1)$ than $Synthetic(0,0)$. The reason might be that an increasing data heterogeneity can make the convergence rate more dependent the delayed value $\tau$.

\textbf{Choice of $\lambda_{0}$.}
Here, we evaluate the choice of $\lambda_{0}$ to the performance of HFL. We set the value of $\lambda_0$ from $0.1$ to $0.9$ on the Fashion MNIST dataset with a fixed communication round $T = 200$, and the results are shown in Figure.~\ref{fig:lambda}. It can be seen that when $\lambda_0 = 0.5,$ HFL has the best the testing accuracy and the training loss, and the testing accuracy results with different $\lambda_0$ ranging from $0.1$ to $0.9$ has a symmetrical pattern with the mean value at $0.5$. Additionally, we examine the largest difference of the testing accuracy in this setting with different values of $\lambda_0$ is only $0.4\%$, which indicates that our algorithm might be adaptive to the setting of $\lambda_0$ during the training process of HFL. 

\textbf{Choice of $\tau$.}
In Sec~\ref{Sec:Convergence}, our discussion suggests that the convergence rate of HFL can be influenced by the choice of $\tau$. We investigate this case on the Fashion MNIST dataset with different maximum $\tau$ values, followed by a fixed communication round $T = 200$. The results in Figure~\ref{fig:tau} come with maximum $\tau \in [10, 100]$ and  $\tau_i, i \in \mathcal{S}_2$ is uniformly distributed. We could notice when $\tau = 10,$ both the testing accuracy and the training loss have the best performance. However, we find an interesting phenomenon that when $\tau = 40, 50$ and $60$, learning performance is significantly better than other settings. We consider the reason of this phenomenon might indicate a non-linear relationship between the setting of $\tau$ and the optimization performance, and this could be a interesting topic for our further research. 

\subsection{Non-convex Optimization Results}
We then evaluate the performance of our HFL algorithm on non-convex optimization problems. The comparison results for HFL against existing FL methods on Sent$140$ and Shakespeare datasets to train a LSTM classifier are provided. The results in Figures~\ref{fig:sent140acc} and \ref{fig:sent140loss} show that our HFL algorithm has the best overall testing accuracy and training loss against existing FL algorithms on Sent140. In particular, the results in Figure~\ref{fig:sent140acc} show that, although HFL has a slower convergence rate compared to FedAvg and FedProx, the testing accuracy is significantly higher. Although there are notable oscillations in the training process of our HFL, we consider it could be tolerated by comparing the amplitude of the oscillation to other FL algorithms. 

The results in Figures~\ref{fig:shakeacc}-\ref{fig:shakeloss} show the comparison on the Shakespeare dataset. From the results, we can see that, compared to the FedAvg and the FedProx, HFL obtains has an improved testing accuracy and training loss. Although the results show a clear gap between HFL and S-SGD method and the convergence rate of HFL is slow in this dataset, HFL still outperforms FedAvg and FedProx with a non-overfitting training loss and at least $12\%$ higher testing accuracy. In conclusion, we believe the proposed HFL algorithm has a better performance against other FL methods for solving non-convex optimization problems.

\section{Conclusion}\label{Sec:Conclusion}
In this paper, we investigated the impact of stragglers on the performance of non-i.i.d. optimization in a heterogeneous network for FL. We proposed a new FL algorithm, called HFL, with two key components: a synchronous kernel and an asynchronous updater to train the joint model under two difference communication scenarios. To incorporate delayed local gradients from stragglers, we designed an adaptive approximation method, called AD-SGD. We demonstrated the effectiveness of HFL through theoretical convergence analysis and experimental evaluations. Theoretically, we provided the convergence guarantee of our HFL on both convex and non-convex optimization problems in heterogeneous network, followed by a discussion on the convergence learning rate. Empirically, we showed that HFL outperforms existing synchronous FL methods on both synthetic and real-world benchmarks.

\newpage
\appendix
\title{Supplementary \\ Stragglers Are Not Disaster: A Hybrid Federated Learning Algorithm with Delayed Gradient}

\maketitle

\section{Related Work}
\textbf{Federated learning.}
Federated Learning (FL) \cite{konevcny2016federated}, is a novel collaborative learning model in a distributed network which is usually with a center server and multiple remote devices. As the development of distributed networks, FL has attracted great interest from the ML research field. In a FL network, a joint model is trained on the server with the dataset distributed among the remote devices. Specially, the training process of the FL model is without data sharing, which protects the data privacy in the network. The joint model is often trained to address distributed optimization problems, e.g., next word prediction \cite{hard2018federated, yang2018applied}. Existing works on the FL mainly focus on the following categories:  i) communication efficiency \cite{konevcny2016federated, mcmahan2017communication, sahu2018federated, smith2017federated, li2019convergence}; ii) distributed optimization problem \cite{sattler2019robust, zhao2018federated, yan2020distributed} and iii) privacy consideration \cite{baruch2019little, fang2019local, bhagoji2019analyzing}. 

\textbf{Heterogeneous optimization in FL.} 
\cite{mcmahan2017communication} first proposes the SOTA Federated Averaging (FedAvg), which provides a convergence guarantee on the i.i.d optimization problems as well as being able to address the communication bottleneck of FL. By training the joint model with only a subset of devices in each communication round, instead of a full device participation scheme, FedAvg significantly increase the training speed of the joint model. Works from \cite{sattler2019robust, zhao2018federated} consider the optimization scenarios with non-i.i.d distributed learning data, however, the results in their works come without the convergence rate. \cite{li2019convergence} investigate the FedAvg algorithm and provides a novel FedAvg mechanism to address non-i.i.d data optimization. A convergence guarantee is provided in \cite{li2019convergence} with two assumptions: i) all the remote devices are active during the training process; ii) the server can randomly access each server in every communication round. To address the FL in a heterogeneous network, \cite{sahu2018federated} proposes FedProx, which is a popular variant of FedAvg with an added quadratic proximal term. This algorithm takes the statistical heterogeneity of the devices into consideration and gives a convergence guarantee on non-convex optimization problems under non-i.i.d settings. However, the existing works, e.g., FedAvg and FedProx, are both developed based on impractical assumptions that can violate the FL settings in real-world applications. 

\section{Convergence Analysis of HFL}

\subsection{Convex Optimization}
To introduce the convergence analysis of HFL against the convex optimization problems, we first introduce several extended assumptions, which have been applied in the previous FL convex optimization researches. In particular, the Assumption~\ref{Assum:Global_1} and \ref{Assum:Global_5} have been made by the previous works \cite{yu2019parallel, stich2018local, stich2018sparsified} and the Assumption~\ref{Assum:Global_3} is widely applied in existing works. 

\begin{assumption}\label{Assum:Global_3}
	\textbf{\emph{(Bounded local objective:})} For the $i$-th device at the $t$-th communication round, we consider the variance of the stochastic gradient against convex optimization problems in this local training process is uniformly bounded as
	\begin{equation}\label{Eq:Var_bound}
	\mathbb{E} ||\nabla F (\mathbf{w}_{t}^{i}, \mathbf{X}_{t}^{i}) ||^2 \leq \sigma^2.
	\end{equation}
\end{assumption}

\begin{assumption}\label{Assum:Global_1}
	\textbf{\emph{($\mu$-Quasi Convex):}} The objective function $F(\cdot)$ for the FL network is differentiable and $\mu$-quasi convex with a constant value $\mu$ that 
	\begin{eqnarray}\label{Eq:Mu_convex}
	&& F(\mathbf{w}^{\star}) \geq F(\mathbf{u}) + \langle \mathbf{w}^{\star}- \mathbf{u}, \nabla F(\mathbf{u})\rangle  + \frac{\mu}{2} {|| \mathbf{w}^{\star} - \mathbf{u}||}^2, 
	\end{eqnarray}
	where $\mathbf{u}$ represents an arbitrary weight status corresponding to the objectives. 
\end{assumption}

\begin{assumption}\label{Assum:Global_5}
\textbf{\emph{$L_2$ smoothness:}} We define the difference between the optimal gradient and the approximation as $\varepsilon$, where $ \varepsilon = ||g(\mathbf{w}_{t}) - g (\mathbf{w}_{t-\tau}) - R(\mathbf{w}_{t-\tau}) (\mathbf{w}_{t} - \mathbf{w}_{t-\tau}) ||$. The difference value $\varepsilon$ is considered to be  $L_2$-smooth with a constant value $L_2$ that
	\begin{equation}
	\varepsilon \leq \frac{L_2}{2} ||\mathbf{w}_{t} - \mathbf{w}_{t-\tau} ||^2,
	\end{equation}
	where $\varepsilon$ can be considered as the higher order item deviation in the Taylor Expansion. 
\end{assumption}

\begin{theorem}\label{Theorem_2} For the convex optimization problems, let all the assumptions in this paper hold that $F(\cdot)$ is $\mu$-convex and $L$-smooth with bounded stochastic gradients, our HFL algorithm satisfies 
\begin{equation}
\mathbb{E} F(\mathbf{w}_{t}) -  F(\mathbf{w}^{\star}) \leq   \frac{ L_2 L^3 \tau^2 G^2 \sigma^2 }{\mu^6 (t+\tau)^2 B_2} +  \frac{L^3 G^2}{2 (t+\tau) \mu^4 B_2}, 
\end{equation}
when $\eta_t \leq \frac{L}{\mu^2tB^4 \Psi_1 \Psi_2}$ and $B_2 = B^8 \Psi_1^2 \Psi_2^2$.
\end{theorem}

\textit{Proof.}
We start the proof of Theorem~\ref{Theorem_2} from an ideal $t$-th communication round. Supposing that at $t$-th round, the server communicates with each remote device, then with a maximum delayed gradient step $\tau$, we have the following relationship
\begin{equation}\label{Eq:convex}
\begin{split}
	 &\mathbb{E} F (\mathbf{w}_{t+\tau+1 }) - F(\mathbf{w}^{\star}) \nonumber\\
	 & \overset{(a_1)}{\leq}  F (\mathbf{w}_{t + \tau}) - F(\mathbf{w}^{\star})  +  \langle  \nabla F(\mathbf{w}_{t+\tau}), \mathbf{w}_{t+\tau+1 } - \mathbf{w}_{t+\tau }\rangle \nonumber + \frac{L}{2} ||\mathbf{w}_{t+\tau+1 } - \mathbf{w}_{t+\tau }||^2 \nonumber \\
& \overset{(a_2)}{\leq}  F (\mathbf{w}_{t + \tau }) - F(\mathbf{w}^{\star}) -  \eta_{t+\tau}  \langle 	 \nabla F(\hat{\mathbf{w}}_{t+\tau }),  \mathbf{w}_{t+\tau+1 } - \mathbf{w}_{t+\tau } \rangle  + \frac{L}{2} ||\mathbf{w}_{t+\tau+1 } - \mathbf{w}_{t+\tau }||^2 \nonumber \\
& \overset{(a_3)}{\leq}  F (\mathbf{w}_{t + \tau}) - F(\mathbf{w}^{\star})  -  \eta_{t+\tau}  \langle \sum_{i \in \mathcal{S}_1} p_i \nabla F_i(\hat{\mathbf{w}}_{t+\tau }), \mathbf{w}_{t+\tau+1 } - \mathbf{w}_{t+\tau } \rangle + \frac{L}{2} ||\mathbf{w}_{t+\tau } - \mathbf{w}_{t+\tau-1 }||^2\nonumber \\
& \overset{(a_4)}{\leq} F (\mathbf{w}_{t + \tau}) - F(\mathbf{w}^{\star}) + \frac{L}{2}||\mathbf{w}_{t+\tau+1} - \mathbf{w}_{t+\tau }||^2 -  \eta_{t+\tau} B^2 \Psi_1 \langle\nabla F(\mathbf{w}_{t+\tau }),  \nabla F(\mathbf{w}_{t+\tau })- \varepsilon \rangle \nonumber \\
& =  F (\mathbf{w}_{t + \tau}) - F(\mathbf{w}^{\star}) + \frac{L}{2}||\mathbf{w}_{t+\tau+1 } - \mathbf{w}_{t+\tau }||^2 -  \eta_{t+\tau} B^4 \Psi_1 \Psi_2 \langle\nabla F(\mathbf{w}_{t+\tau}), \nabla F(\mathbf{w}_{t+\tau })- \varepsilon \rangle,  \nonumber
\end{split}
\end{equation}
where the inequality $(a_1)$ comes from the $L$-smooth assumption, and we consider the joint model is updated via the virtual synchronous sequence $\hat{\mathbf{w}}$ in $(a_2)$. Additionally, at $(a_3),$ we have the results from $\hat{\mathbf{w}}_{t} = \sum_{i \in \mathcal{S}_1}  p_i \mathbf{w}_{t}^{i}.$ And from the $L_2$ smoothness in the Assumption~C, we represent the expansion of $g(\mathbf{w}_{t+\tau}) = \mathbf{w}_{t+\tau +1} - \mathbf{w}_{t+\tau}$. Then, let $A_1 =\langle\nabla F(\mathbf{w}_{t+\tau }), \nabla F(\mathbf{w}_{t+\tau})-  \varepsilon \rangle$, we have 
\begin{equation}
	A_1 \leq \langle\nabla F(\mathbf{w}_{t+\tau }), \nabla F(\mathbf{w}_{t+\tau}) \rangle - \langle \nabla F(\mathbf{w}_{t+\tau }),  \varepsilon \rangle, \nonumber
\end{equation}
from the $\mu$-strongly convex assumption we could have 
\begin{equation}
    - \langle\nabla F(\mathbf{w}_{t+\tau }), \nabla F(\mathbf{w}_{t+\tau }) \rangle \leq \frac{-2\mu^2}{L} (F(\mathbf{w}_{t+\tau }) - F(\mathbf{w}^{\star})).
\end{equation}
And according to the Cauchy–Schwarz inequality easily we could have  
\begin{equation}
\begin{split}
\eta_{t+ \tau} \langle \nabla F(\mathbf{w}_{t+\tau }),  \varepsilon \rangle & \leq \eta_{t+ \tau}  ||\nabla F(\mathbf{w}_{t+\tau })|| \varepsilon \nonumber\\
  & \leq \eta_{t+ \tau} \sigma^2 \frac{L_2}{2} ||\mathbf{w}_{t+\tau} - \mathbf{w}_{t} ||^2 \leq  \eta_{t+ \tau} \sigma^2 G^2 \frac{L_2}{2} \tau \sum_{j=0}^{\tau-1} \eta_{t+ j}^2, \nonumber
\end{split}
\end{equation}
for this inequality, when $\eta_t \leq \frac{L}{\mu^2tB^4 \Psi_1 \Psi_2}$, we could have $\sum_{j=0}^{\tau-1} \eta_{t+ j}^2 \leq \frac{L^2 \tau}{\mu^4 t(t+\tau)B^8 \Psi_1^2 \Psi_2^2} \leq \frac{2L^2 \tau}{\mu^4 (t+\tau)^2B^8 \Psi_1^2 \Psi_2^2}.$

Thus,  we get back to $\mathbb{E} F (\mathbf{w}_{t+\tau+1 }) - F(\mathbf{w}^{\star}) $ and have 
\begin{equation}
	\mathbb{E} F (\mathbf{w}_{t+\tau+1 }) - F(\mathbf{w}^{\star}) \leq  (1 - \frac{2}{t + \tau } ) \mathbb{E} F (\mathbf{w}_{t+\tau }) - F(\mathbf{w}^{\star}) + \frac{L_2 L^3 \tau^2 G^2 \sigma^2}{\mu^6 (t + \tau)^3 B^8 \Psi_1^2 \Psi_2^2} + \frac{L^3 G^2}{2 (t + \tau)^2 \mu^4  B^8 \Psi_1^2 \Psi_2^2}. \nonumber
\end{equation}
Let $B_2 = B^8 \Psi_1^2 \Psi_2^2$ and rearrange the result we could get 
\begin{equation}
\begin{split}
\mathbb{E} F(\mathbf{w}_{t}) -  F(\mathbf{w}^{\star}) 
&\leq  \sum_{j = t}^{t+\tau} \frac{L_2 L^3 \tau^2 G^2 \sigma^2 }{\mu^6 (t+j)^3 B_2} +  \frac{L^3 G^2}{2 (t+j)^2 \mu^4 B_2} \nonumber\\
&\leq  (t+\tau)\left( \frac{L_2 L^3 \tau^2 G^2 \sigma^2 }{\mu^6 (t+\tau)^3 B_2} +  \frac{L^3 G^2}{2 (t+\tau)^2 \mu^4 B_2} \right) \nonumber\\
&\leq   \frac{ L_2 L^3 \tau^2 G^2 \sigma^2 }{\mu^6 (t+\tau)^2 B_2} +  \frac{L^3 G^2}{2 (t+\tau) \mu^4 B_2}. \nonumber
\end{split}
\end{equation}

\textit{Proof done.}

\begin{corollary}
\textbf{\emph{(Convergence rate: convex):}} 
Following the above proof steps, we provide the convergence guarantee of the convex optimization problems in our proposed HFL algorithm. In particular, when $\eta_t$ is bounded by $\frac{L}{\mu^2tB^4 \Psi_1 \Psi_2}$, the optimization bound of HFL could be represented as two parts: a high-order part $\frac{ L_2 L^3 \tau^2 G^2 \sigma^2 }{\mu^6 (t+\tau)^2 B_2}$ and a low-order part $\frac{L^3G^2}{2(t+\tau) \mu^4 B_2}$.

Note that the high-order part would converge to a stationary point faster than the low-order part. Thus, we introduce the convergence rate of our HFL which follows the low-order part that 
\begin{equation}
    \mathbb{E} F(\mathbf{w}_{t}) -  F(\mathbf{w}^{\star}) \leq  \mathcal{O}(\frac{1}{t+\tau}).
\end{equation}
\end{corollary}

\subsection{Non-convex Optimization}

\begin{theorem}\label{Theorem_1}
For the non-convex problems under the Assumption 1-2, we consider the model convergence with a constant learning rate $\eta_t$ that
\begin{equation}
	\min_{t \in T} \mathbb{E}|| \nabla F(\mathbf{w}_t)|| ^2 
	 \leq  \frac{1}{T\eta_t B_1} \left[ F(\mathbf{w}_{0}) -  F(\mathbf{w}^{\star}) ) \right],
\end{equation}
\end{theorem}
where $T$ is considered to be a bound of $\tau$ that $t+ \tau \leq T$, and $B_1 = B^4 \Psi_1 \Psi_2$. Specifically, the bound of $\eta_t$ needs to satisfy the following inequality 
\begin{equation}\label{Eq:condition}
    \frac{\eta_{t}^2 LG^2 \sqrt{\frac{\Psi_2 B_1}{\Psi_1}}}{2} - \eta_{t}B_1G^2 \leq 0.
\end{equation}

\textit{Proof.} Following the $L$-smooth assumption for the joint objective, we start from the result at the $(t + \tau + 1) $-th step as

\begin{equation}
\begin{split}
 \mathbb{E} F(\mathbf{w}_{t+\tau + 1}) & -  F(\mathbf{w}_{t+\tau})\\
& \overset{(b_1)}{\leq}  \langle\nabla F(\hat{\mathbf{w}}_{t+\tau}), \mathbf{w}_{t+\tau + 1} - \mathbf{w}_{t+\tau}\rangle + \frac{L}{2} ||\mathbf{w}_{t+\tau + 1} - \mathbf{w}_{t+\tau}||^2 \nonumber\\
& \overset{(b_2)}{\leq}  - \eta_{t+\tau} \langle\nabla F(\hat{\mathbf{w}}_{t+\tau}),  R(\mathbf{w}_{t}) + g(\mathbf{w}_{t})\rangle + \frac{L}{2} ||\mathbf{w}_{t+\tau + 1} - \mathbf{w}_{t+\tau}||^2 \nonumber\\
& \overset{(b_3)}{\leq}   - \eta_{t+\tau} \langle\sum_{i \in \mathcal{S}_1} p_i \nabla F_i(\mathbf{w}_{t+\tau}), \sum_{j \in \mathcal{S}_2}g(\mathbf{w}_{t+\tau}^{j})\rangle + \frac{L}{2} ||\mathbf{w}_{t+\tau + 1} - \mathbf{w}_{t+\tau}||^2 \nonumber\\
&\leq - \eta_{t+\tau}B^4 \Psi_1 \Psi_2 \langle \nabla F(\mathbf{w}_{t+\tau}), g(\mathbf{w}_{t+\tau}) \rangle + \frac{L}{2} ||\mathbf{w}_{t+\tau + 1} - \mathbf{w}_{t+\tau}||^2 \nonumber\\
& \overset{(b_4)}{\leq} - \eta_{t+\tau}B^4 \Psi_1 \Psi_2( G^2 + || \nabla F(\mathbf{w}_{t+\tau})||^2 ) + \frac{\eta_{t+\tau}^2 L B^2 \Psi_2}{2}||g(\mathbf{w}_{t+\tau})||^2.  \nonumber\\
\end{split}
\end{equation}
The derivations of the inequality $(b_1)$, $(b_2)$ and $(b_3)$ follow the same steps in \eqref{Eq:convex}. And the results in $(b_4)$ come from the feature of the inner product vector: for two vectors $\mathbf{u}, \mathbf{v} \in \mathbb{R},$ we have $\langle \mathbf{u}, \mathbf{v} \rangle \leq ||\mathbf{u}||^2 ||\mathbf{v}||^2.$  Then, let $B_1 = B^4 \Psi_1 \Psi_2  $, we have
\begin{equation}
\mathbb{E} F(\mathbf{w}_{t+\tau + 1}) -  F(\mathbf{w}_{t+\tau}) \leq - \eta_{t+\tau}B_1 \mathbb{E} || \nabla F(\mathbf{w}_{t+\tau})||^2 + \frac{\eta_{t+\tau}^2 LG^2 \sqrt{\frac{\Psi_2 B_1}{\Psi_1}}}{2} - \eta_{t+\tau}B_1G^2. \nonumber
\end{equation}

Then, we could summarizing the previous inequality from $t=1$ to $T = t + \tau$ that 
\begin{equation}
\mathbb{E} F(\mathbf{w}_{T + 1}) -  F(\mathbf{w}_{1})\leq - \eta_{t}B_1 \sum_{t=1}^{T}  \mathbb{E} || \nabla F(\mathbf{w}_{t})||^2 + \Phi,  
\end{equation}
where $\Phi \triangleq   \frac{T\eta_{t}^2 LG^2 \sqrt{\frac{\Psi_2 B_1}{\Psi_1}}}{2} - \eta_{t}B_1G^2.$ Thus, when the value of $\Phi \leq 0,$ the (9) comes with the convergence guarantee. By replacing the joint model $\mathbf{w}$ with the optimal result $\mathbf{w}^{\star}$ we could have 
\begin{equation}
 \frac{1}{T} \sum_{t=1}^{T} \mathbb{E} || \nabla F(\mathbf{w}_{t})||^2 
\leq  \frac{1}{T \eta_{t} B_1}  F(\mathbf{w}_{t}) -  F(\mathbf{w}^{\star}) .
\end{equation}
Then the proof of convergence is done. 

\begin{corollary}
\textbf{\emph{(Convergence rate: non-convex):}} Let the learning rate is bounded as $\eta_t \leq \frac{2}{L} \sqrt{\frac{\Psi_1 B_1}{\Psi_2}},$ then we have the value $\Phi \leq 0$ by satisfying the inequality in Eq.~\eqref{Eq:condition}. Thus, we obtain the convergence rate for the HFL algorithm on  non-convex problems as 
\begin{equation}\label{Eq:rate_non-convex2}
    \min_{t \in T} \mathbb{E} || \nabla F(\mathbf{w}_t)|| ^2  \leq \mathcal{O}(\frac{1}{T}),
\end{equation}
\end{corollary}
where $T \geq t + \tau$, in this condition, we obtain a convergence rate of our HFL algorithm with the maximum delayed gradient $\tau$.

\clearpage
\section{Experiments}

\subsection{Detailed Experimental Setting}
\textbf{Implementation.}
In this paper, we evaluate our HFL algorithm on multiple tasks, models and datasets in a simulated federated network. All the experiments are performed with Pytorch \cite{paszke2017automatic} platform at version $1.6.0$, and we represent the remote devices by lightweight threads with Python threading library. In order to simulate the asynchronous update process, we assign $\tau$-related flags to difference remote device threads.  

\textbf{Models.} For the convex optimization problems, we evaluate our algorithm with a multinomial logistic regression model. We represent the prediction model as $f(\mathbf{w}; \mathbf{x}^{i})$ where $\mathbf{w} = (\mathbf{W}, \mathbf{b})$, which satisfies $f(\mathbf{w}; \mathbf{x}^{i}) = \softmax(\mathbf{W} \mathbf{x}^{i} + \mathbf{b})$. Then we have the loss function as 
\begin{equation}
	\frac{1}{N} \sum_{i=1}^{N} \crossentropy(f(\mathbf{w}; \mathbf{x}^{i})) + \epsilon ||\mathbf{w}||^2,   \nonumber
\end{equation}
where we define $\epsilon = 10^{-4}$ in this paper. And for the non-convex optimization problems, in this paper we introduce a LSTM classifier with a recurrent neural network (RNN) \cite{zaremba2014recurrent} architecture. 

\textbf{Datasets.} In this part, we provide the full introduction of the datasets and the numerical information of datasets is summarized in Table~\ref{Table:dataset}. 

\textbf{1) Synthetic Data.} The synthetic data in this paper are developed followed by the original setup from the work in \cite{shamir2014communication, li2019convergence}, which is designed to simulate a quadratic problem. Specifically, for the $i$-th remote device, we generate the learning data samples $(\mathbf{x}^{i}, {y}^{i})$ from a softmax function ${y}^i = \arg\max (softmax(\mathbf{w}^{i} \mathbf{x}^{i} + \mathbf{b}^{i}))$, where we define $\mathbf{x}^{i} \in \mathbb{R}^{60}$, $\mathbf{w}^{i} \in \mathbb{R}^{60 \times 10}$ and $\mathbf{b}^{i} \in \mathbb{R}^{10}$. Additionally, we define the distribution of $\mathbf{w}^{i}, \mathbf{b}^{i}$ separately as  $\mathbf{w}^{i} \sim \mathcal{N}(u^i, 1)$ and $\mathbf{b}^{i} \sim \mathcal{N}(u^i, 1)$, where we consider $u^i \sim \mathcal{N}(0, \gamma)$. For $\mathbf{x}^{i},$ we consider $\mathbf{x}^{i} \sim \mathcal{N}(v^i, \Sigma)$, where $\Sigma$ represents a diagonal covariance matrix and $v^i \sim \mathcal{N}(B^i, 1)$, $B^i \sim \mathcal{N}(0, \xi)$. In this situation, we use the parameters  $\gamma$ and $\xi$ to manipulate the heterogeneity of the synthetic dataset. 

\textbf{2) Fashion MNIST \cite{xiao2017/online}.} For the real dataset, we introduce Fashion MNIST in this paper as it has been popular among the ML research filed in the recent years. In order to build a non-i.i.d case, we consider each remote device only contains two labels of learning samples and to show the heterogeneity, we distribute the learning data from a power law. 

\textbf{3) Shakespeare \cite{mcmahan2017communication}.} The Shakespeare dataset is developed from ``The Complete Works of William Shakespeare \cite{mcmahan2017communication}". The task of this dataset is to predict the next-character from a input sequence. The number of characters(classes) is $80$ and the total number of input sequence is $517, 106$.

\textbf{4) Sent140 \cite{go2009twitter}.} Sent140 is a text sentiment analysis dataset and popular in the non-convex problem experimental settings in the ML field. Sent140 provides a task to find to corresponding twitter account with a sequence of $25$ characters as input. 

\textbf{5) MNIST \cite{lecun1998gradient}.} MNIST is a classic handwritten digits dataset from $0$-$9$ for image classification problems. The data participation of MNIST in this paper follows the same setups in the Fashion MNIST and we use this dataset for extended evaluations in the supplementary material.

\begin{table}[t!]
	\centering
	\caption{Description of the datasets in this paper.}
	\begin{tabular}{|c|c|c|c|c|}
		\hline
		\textbf{Dataset}& \textbf{Size} & \textbf{Classes} & \textbf{Devices}   \\
		\hline
		{Fashion MNIST}& $70,000$ & $10$ &  $100$  \\
		\hline
		{$Synthetic(0,0)$}& $42,522$ & $10$  &  $100$  \\
		\hline
		{$Synthetic(0.5,0.5)$}& $42,522$ & $10$  &  $100$  \\
		\hline
		{$Synthetic(1,1)$}& $27,348$ & $10$  &  $100$  \\
		\hline
		{Shakespeare}& $517,106$ & $80$  &  $100$  \\
		\hline
		{Sent140}& $40783$ & null  &  $100$ \\
		\hline
		{MNIST}& $70,000$ & $10$ &  $100$  \\	
		\hline
	\end{tabular}
	\label{Table:dataset}
\end{table}

\subsection{Results for Additional Experiments}

\begin{figure}[t!]
    \centering
    \begin{subfigure}{0.32\columnwidth}
        \includegraphics[width = 1\columnwidth]{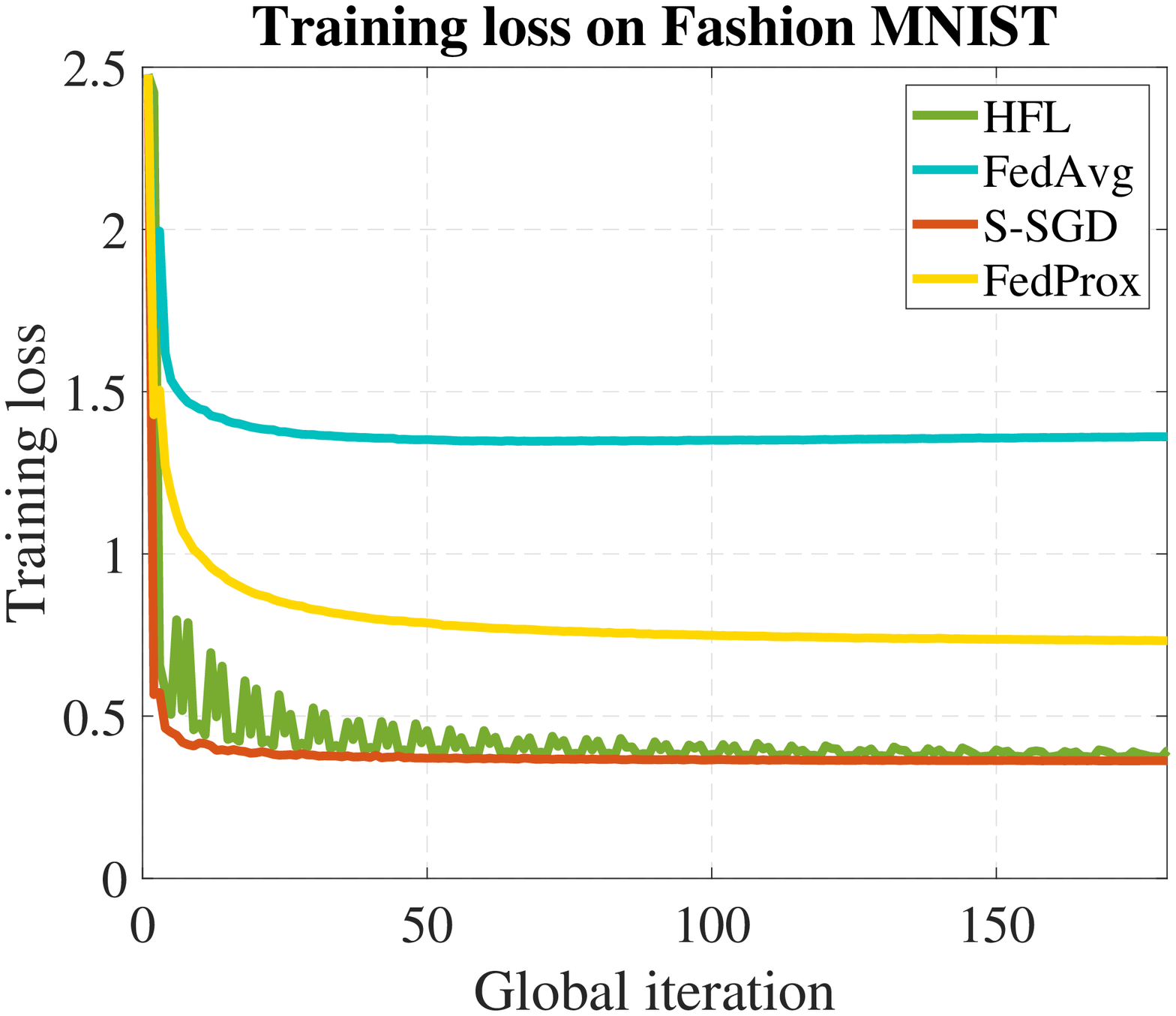}
        \caption{}
        \label{fig:fmnistloss}
    \end{subfigure}
    \begin{subfigure}{0.32\columnwidth}
        \includegraphics[width = 1\columnwidth]{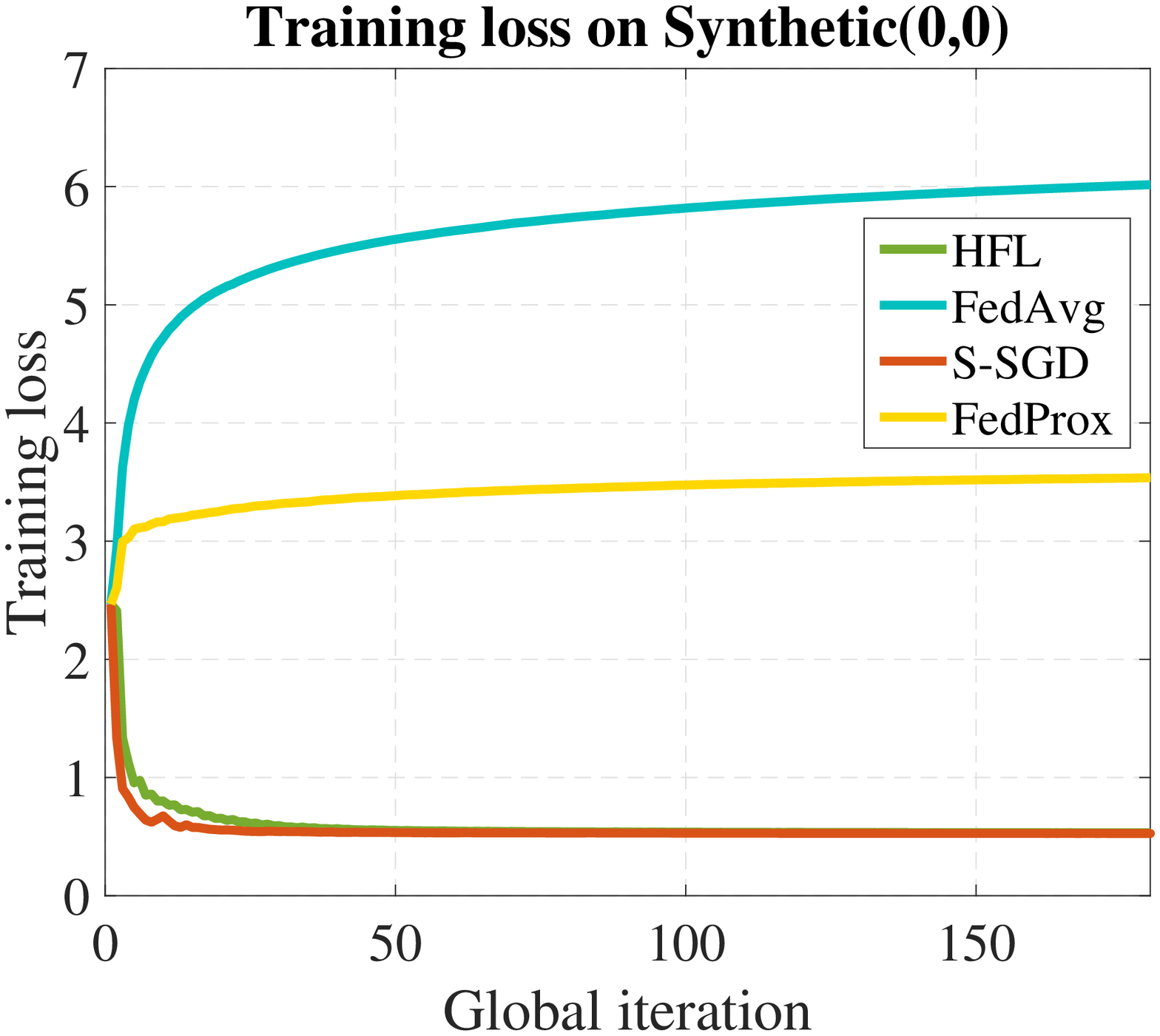}
        \caption{}
        \label{fig:synthetic00noniid_loss}
    \end{subfigure}
        \begin{subfigure}{0.32\columnwidth}
        \includegraphics[width = 1\columnwidth]{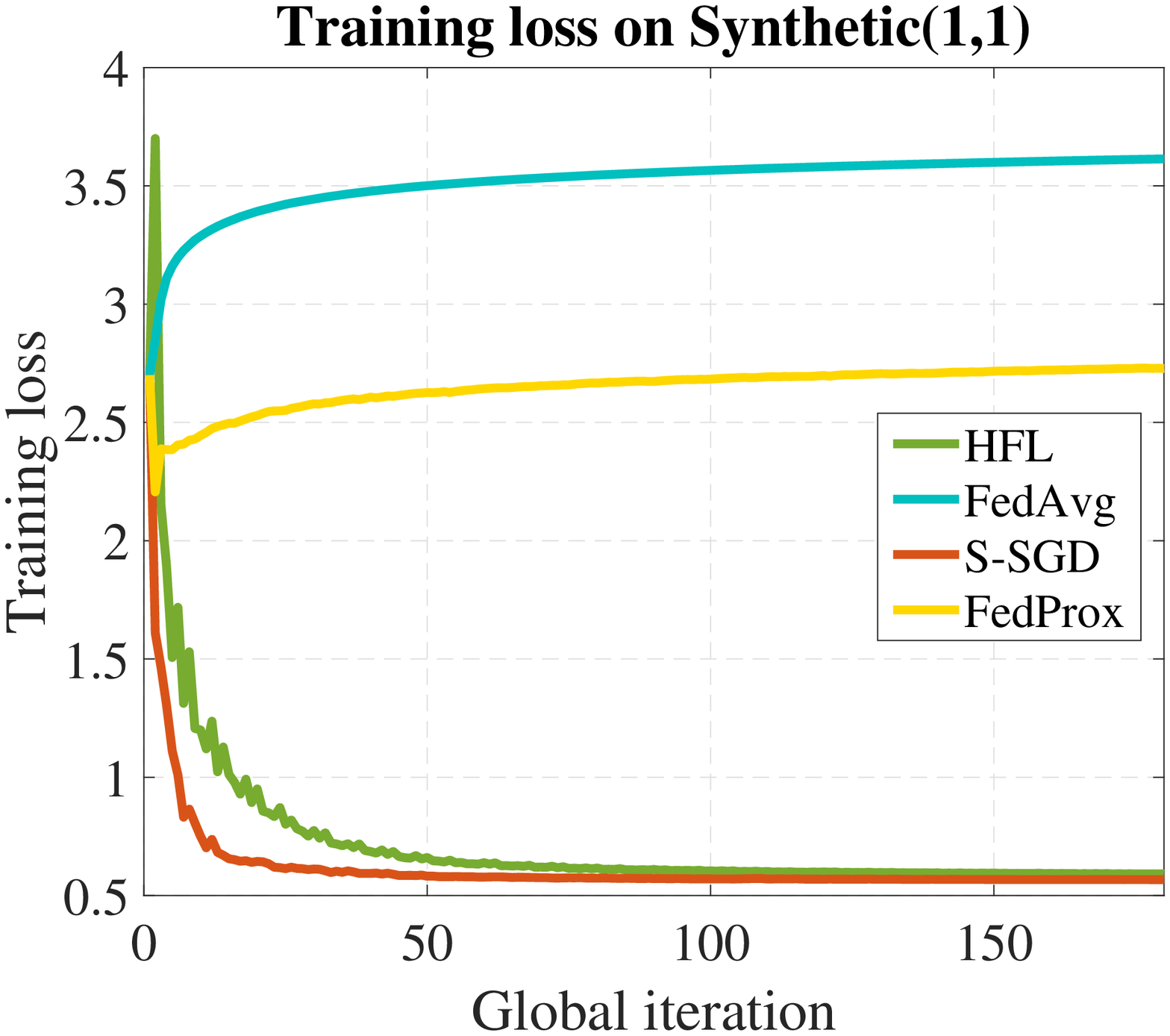}
        \caption{}
        \label{fig:synthetic11noniid_loss}
    \end{subfigure}
    \caption{The learning performance comparison of our HFL to the excising benchmarks on convex problems with non-i.i.d distributed datasets: a) training loss of Fashion MNIST; b) training loss of $Synthetic(0,0)$; c) training loss of $Synthetic(1,1)$;}
    \label{fig:loss}
\end{figure}

\textbf{Training loss of experiments in the paper.}
We first introduce the training loss results in Figure~\ref{fig:loss}. Specially, the training loss experimental settings follow the testing accuracy evaluation in Figure~2. We consider the results of training loss support the analysis of comparison between HFL against existing benchmarks in Sec.~5 in the main paper. 

\begin{figure}[t!]
    \centering
        \begin{subfigure}{0.49\columnwidth}
        \includegraphics[width = 1\columnwidth]{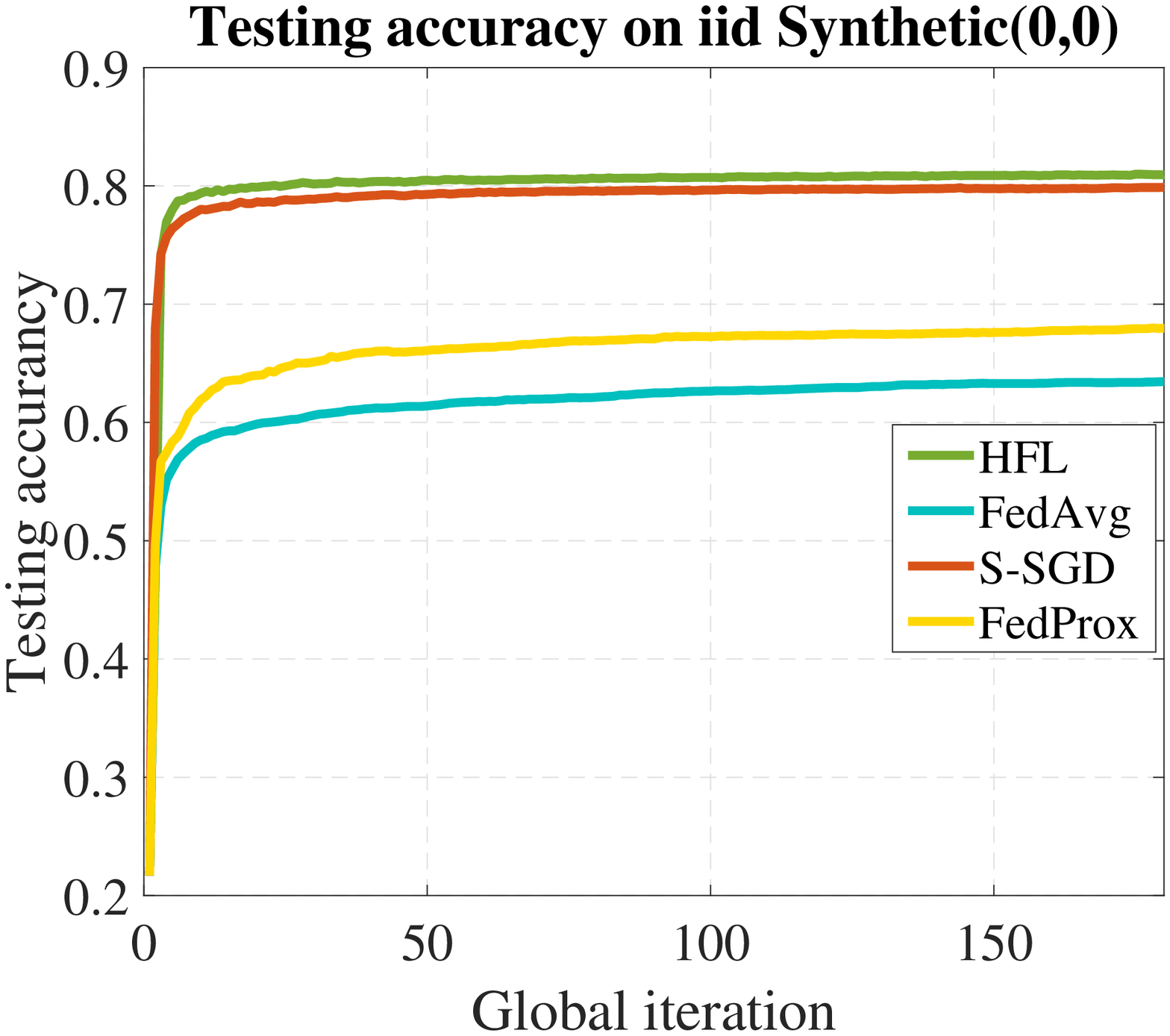}
        \caption{}
        \label{fig:syntheticiid_00_acc}
    \end{subfigure}
        \begin{subfigure}{0.49\columnwidth}
        \includegraphics[width = 1\columnwidth]{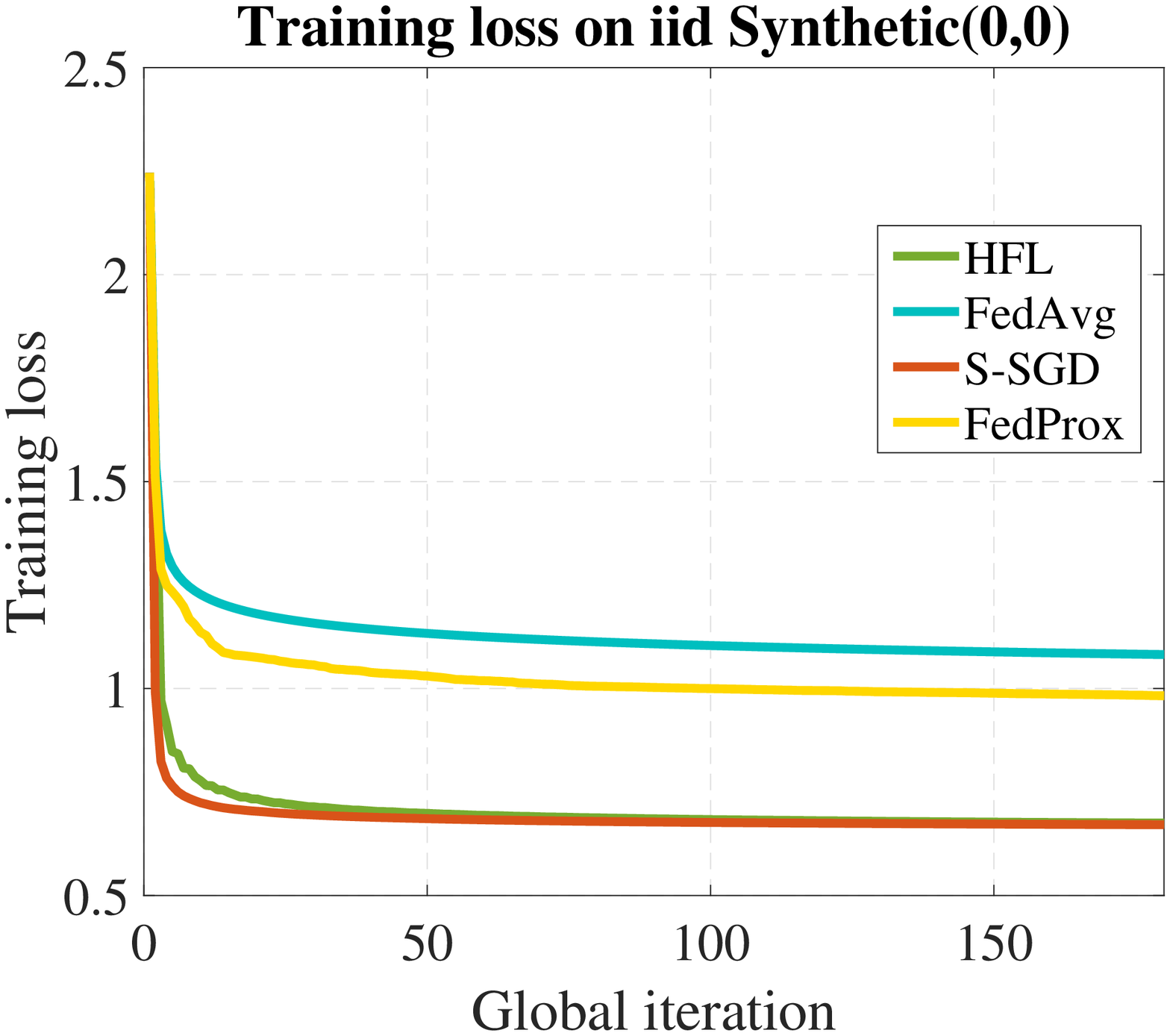}
        \caption{}
        \label{fig:syntheticiid_00_loss}
    \end{subfigure}
    \caption{The learning performance for the compared benchmarks on the i.i.d distributed dataset $Synthetic(0,0)$: a) the testing accuracy; b) the training loss.}
    \label{fig:synthetic00}
\end{figure}

\textbf{Extended experiments with i.i.d datasets.} 
Then, we evaluate the performance of our proposed HFL algorithm against the compared benchmarks on the i.i.d optimization problems in an identically distributed $Synthetic(0,0)$ dataset. The results are shown in Figure~\ref{fig:synthetic00} which contains the training loss and the testing accuracy. 

\begin{figure}[t!]
    \centering
        \begin{subfigure}{0.49\columnwidth}
        \includegraphics[width = 1\columnwidth]{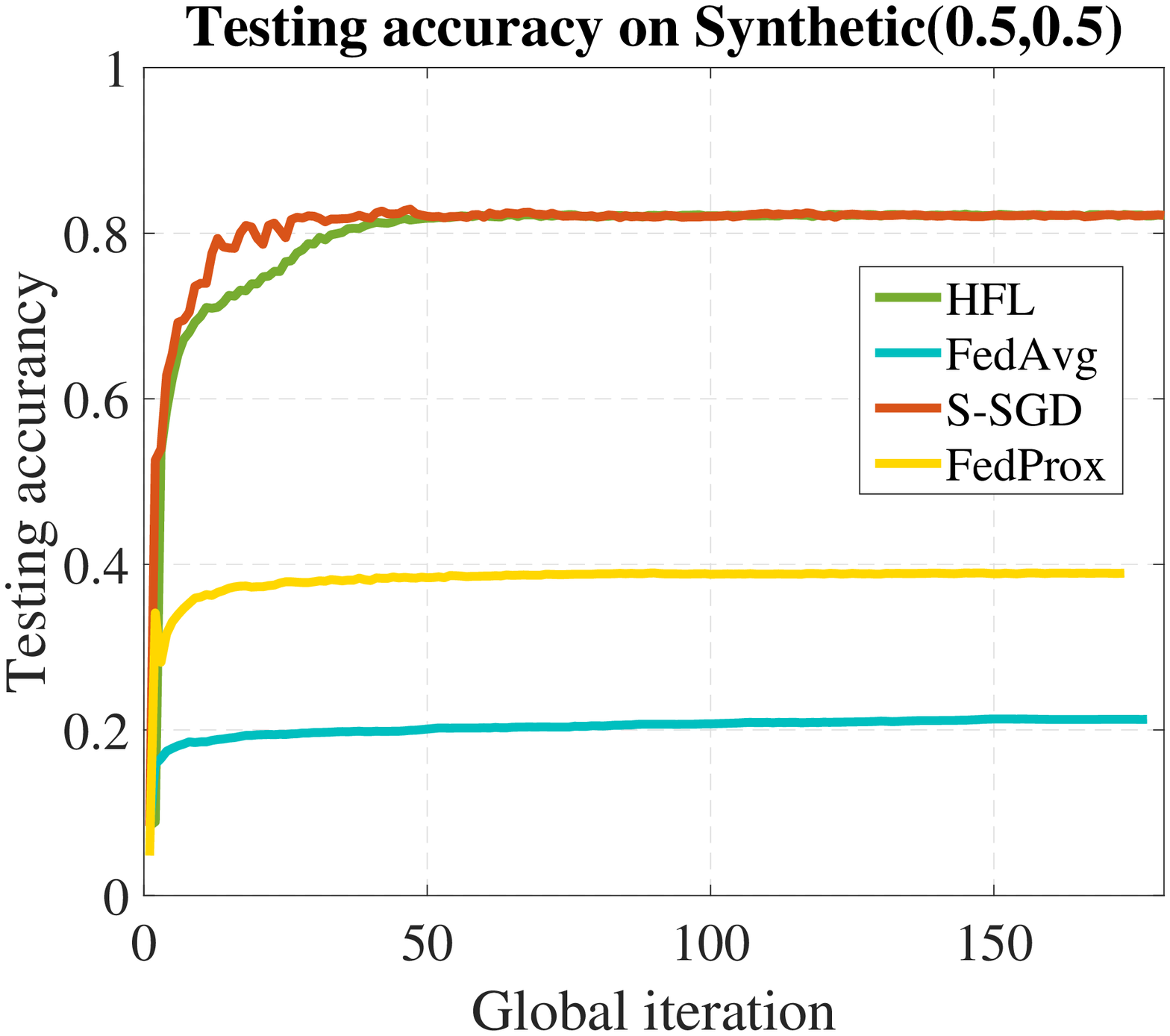}
        \caption{}
        \label{fig:synthetic_0505_acc}
    \end{subfigure}
        \begin{subfigure}{0.49\columnwidth}
        \includegraphics[width = 1\columnwidth]{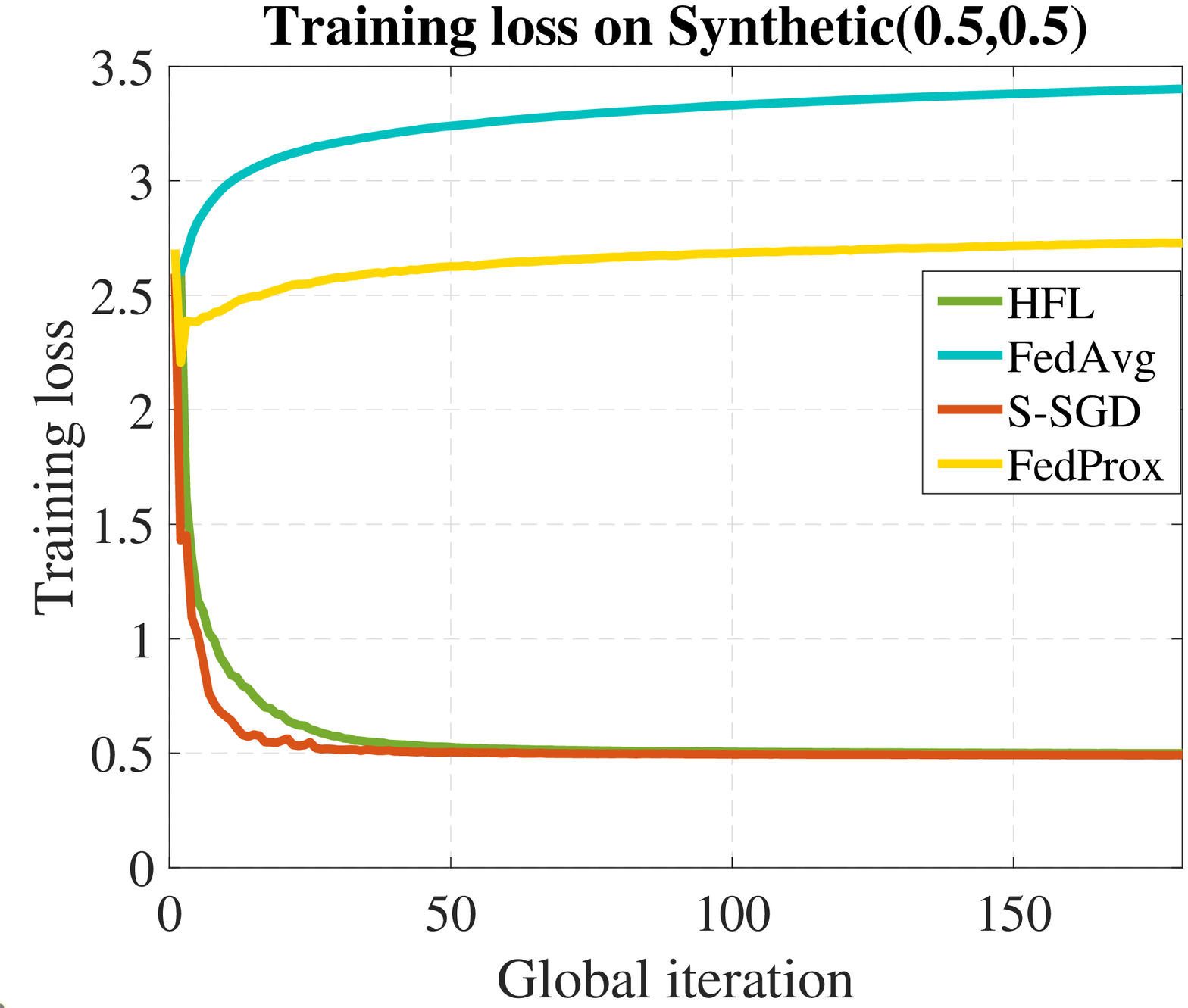}
        \caption{}
        \label{fig:synthetic_0505_loss}
    \end{subfigure}
    \caption{The learning performance for the compared benchmarks on the non-i.i.d distributed dataset $Synthetic(0.5,0.5)$: a) the testing accuracy; b) the training loss.}
    \label{fig:synthetic0505}
\end{figure}

\begin{figure}[t!]
    \centering
        \begin{subfigure}{0.49\columnwidth}
        \includegraphics[width = 1\columnwidth]{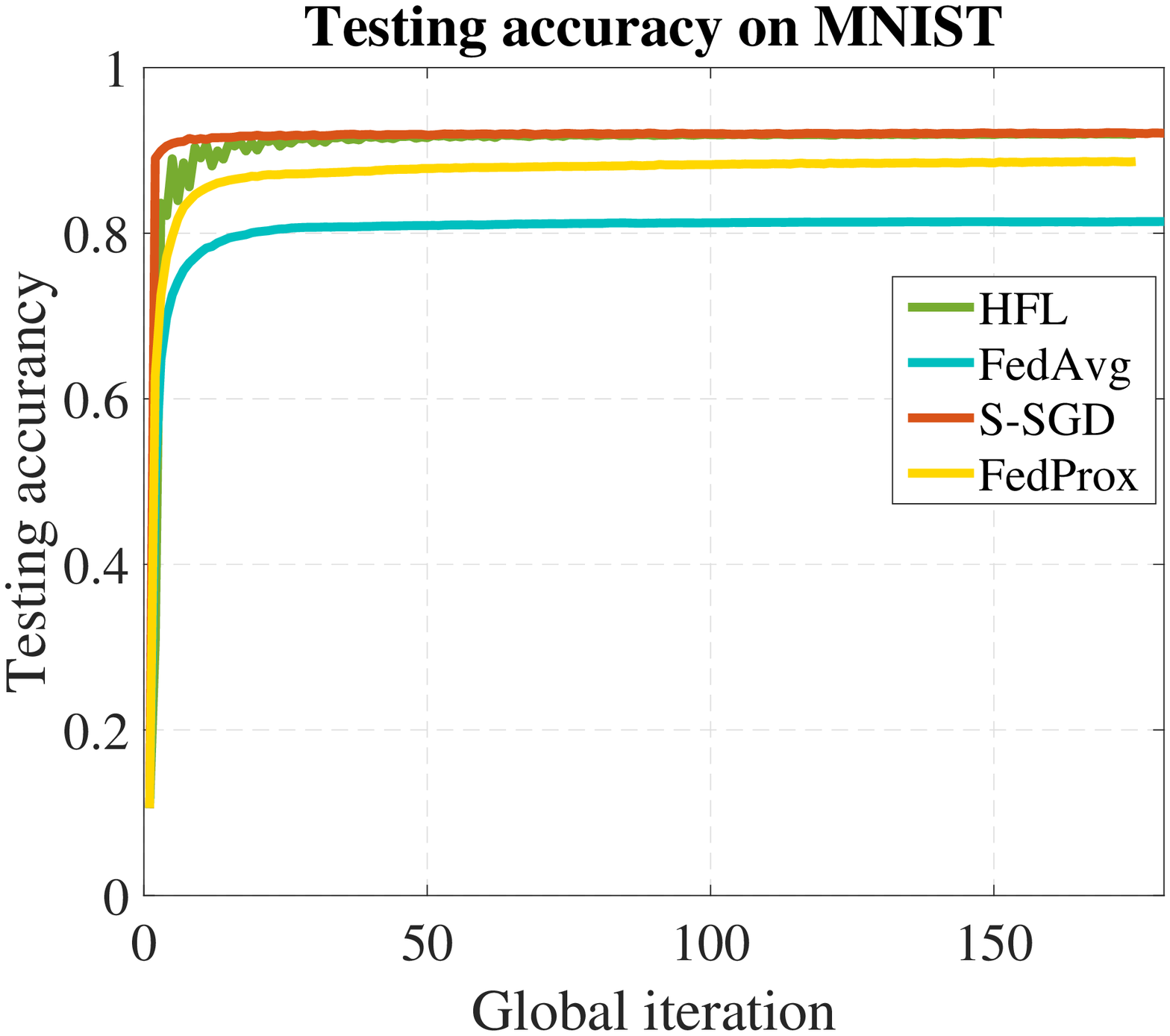}
        \caption{}
        \label{fig:mnist_acc}
    \end{subfigure}
        \begin{subfigure}{0.49\columnwidth}
        \includegraphics[width = 1\columnwidth]{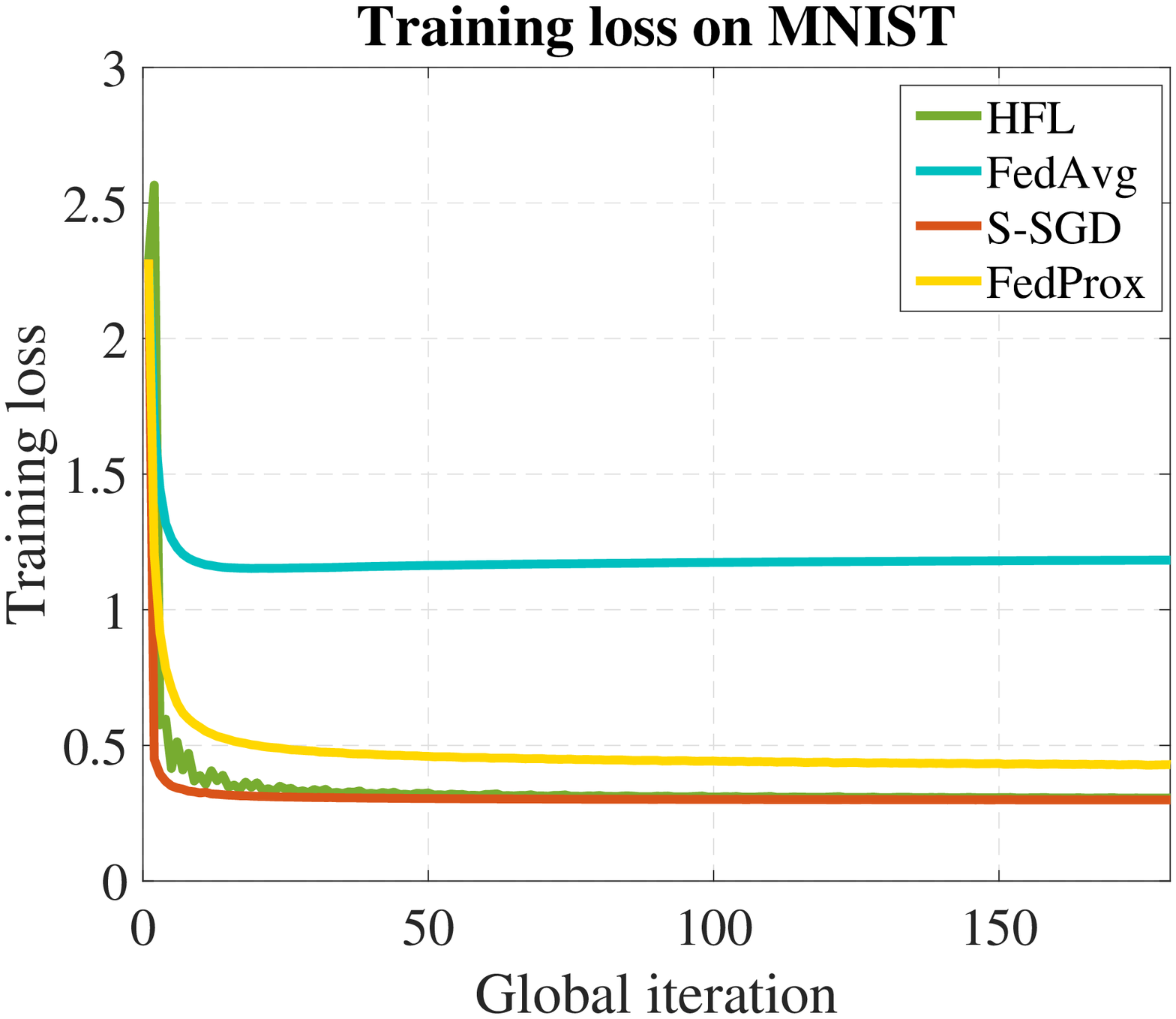}
        \caption{}
        \label{fig:mnist_loss}
    \end{subfigure}
    \caption{The learning performance for the compared benchmarks on the non-i.i.d distributed MNIST dataset: a) the testing accuracy; b) the training loss.}
    \label{fig:mnist}
\end{figure}

\textbf{Extended experiments with non-i.i.d datasets.} 
We conduct extended experimental evaluation for HFL algorithm against the benchmarks under the non-i.i.d distributed training datasets. We first show the evaluation on a non-i.i.d distributed $Synthetic(0.5, 0.5)$ dataset, the results are shown in Figure~\ref{fig:synthetic0505}. Additionally, we introduce the experimental results on the non-i.i.d distributed MNIST dataset, the compared testing accuracy and training loss for the HFL and the benchmarks are shown in Figure~\ref{fig:mnist}.

\begin{figure}[t!]
    \centering
        \begin{subfigure}{0.49\columnwidth}
        \includegraphics[width = 1\columnwidth]{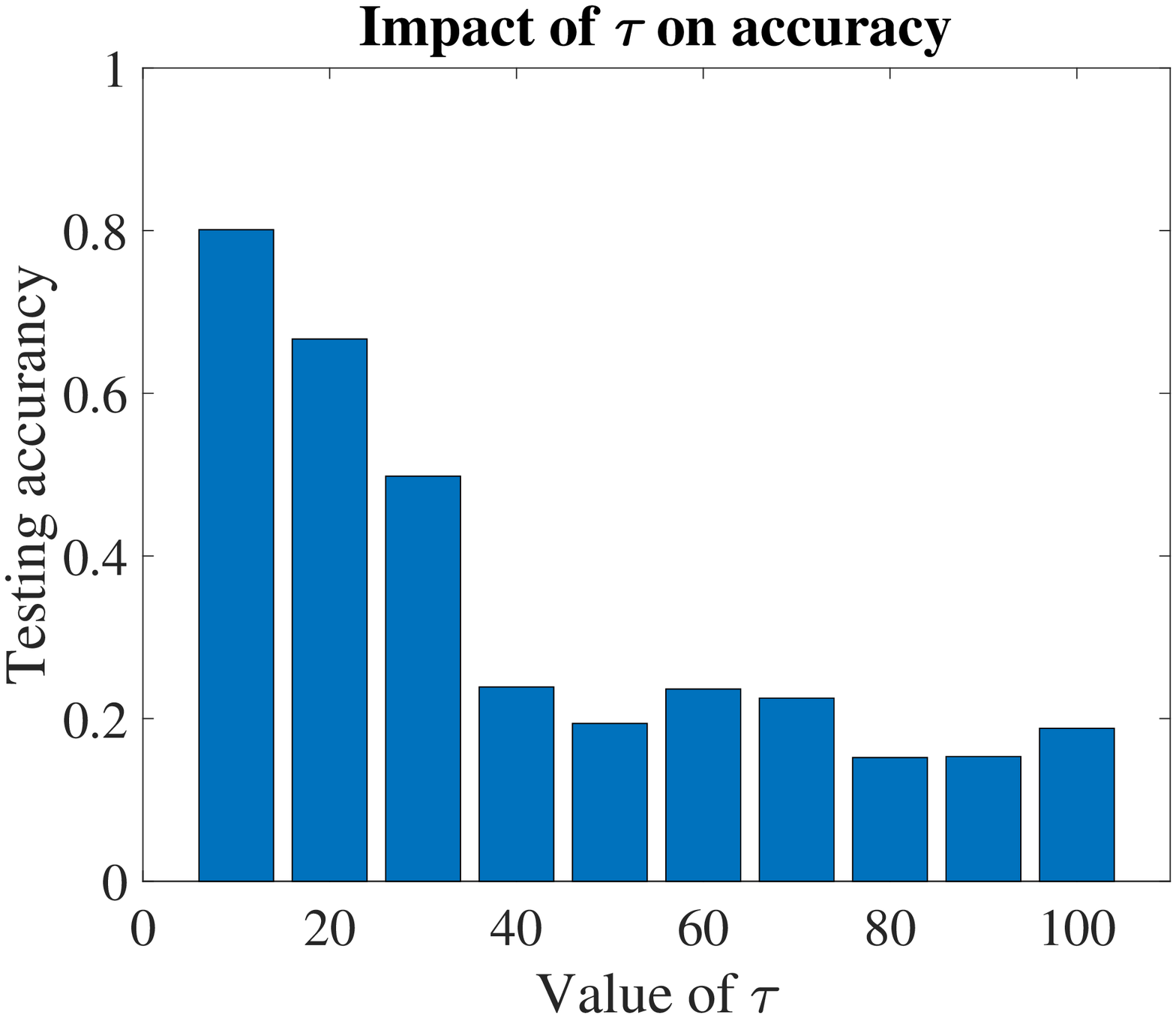}
        \caption{}
        \label{fig:synthetic11_tau_acc}
    \end{subfigure}
        \begin{subfigure}{0.49\columnwidth}
        \includegraphics[width = 1\columnwidth]{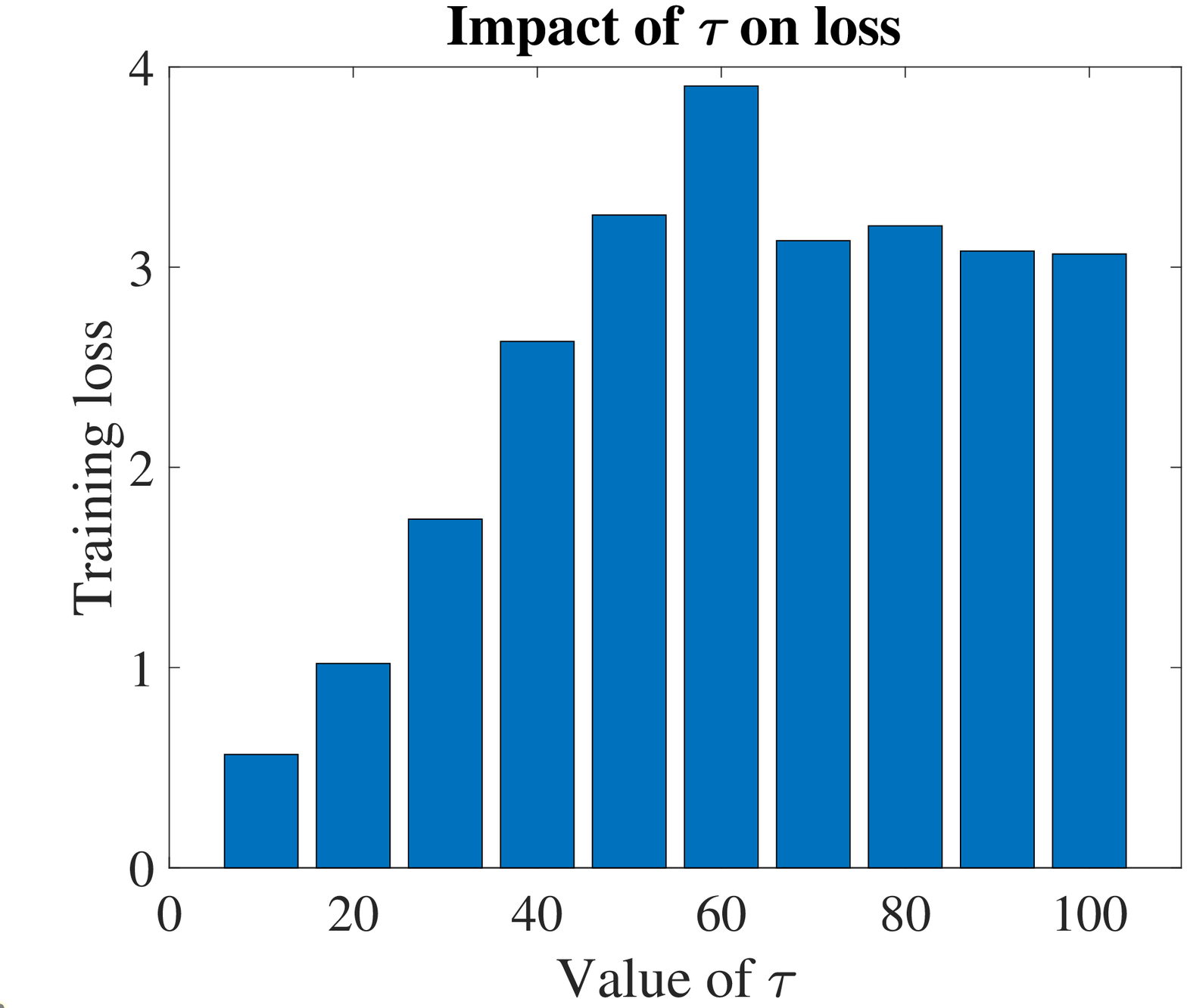}
        \caption{}
        \label{fig:synthetic11_tau_loss}
    \end{subfigure}
    \caption{Choice of $\tau$ on the non-i.i.d distributed $Synthetic(1,1)$ dataset: a) the learning accuracy of HFL with different maximum $\tau$ values; b) the training loss of HFL with different maximum $\tau$ values.}
    \label{fig:tau_11}
\end{figure}

\begin{figure}[t!]
    \centering
        \begin{subfigure}{0.49\columnwidth}
        \includegraphics[width = 1\columnwidth]{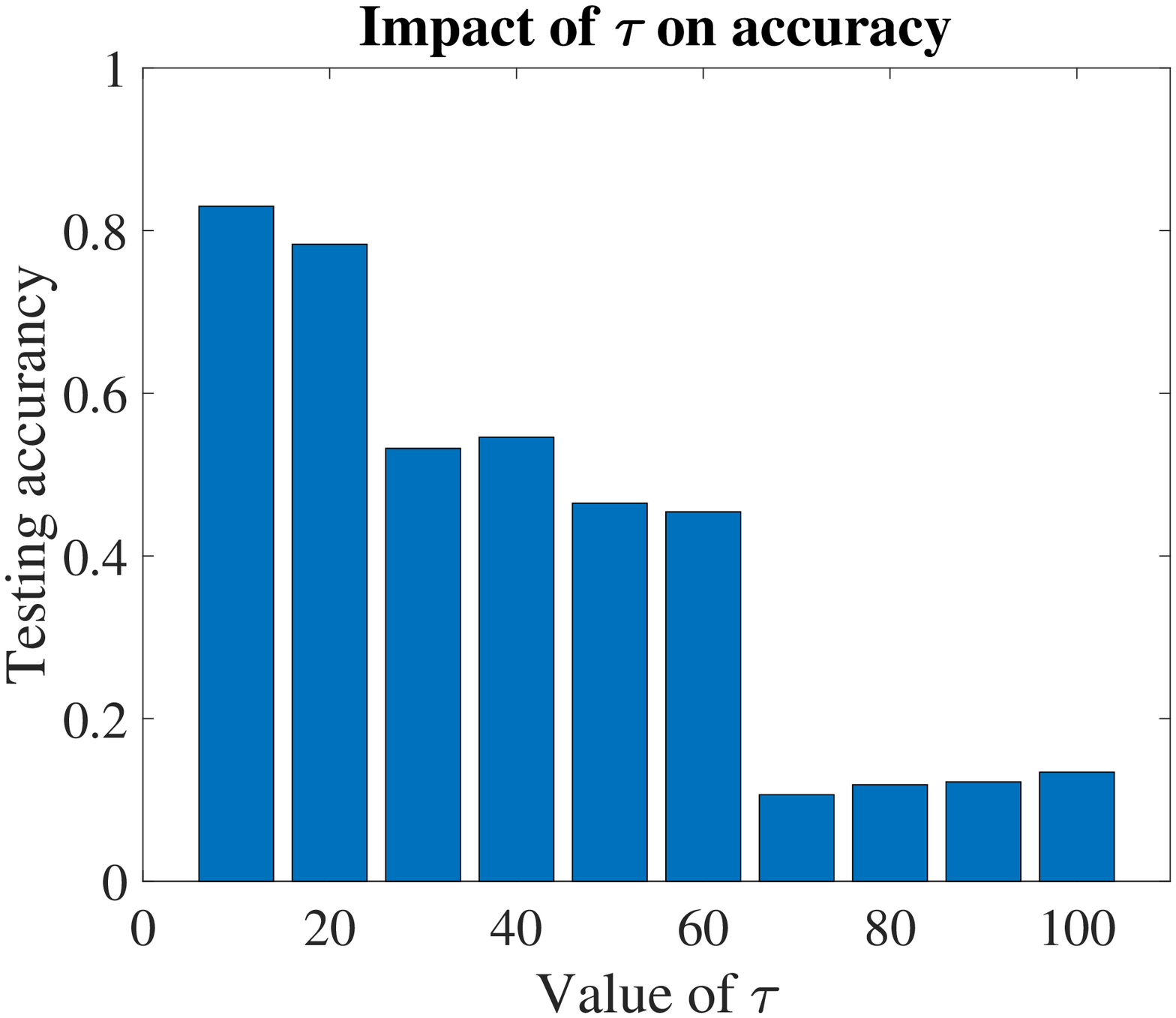}
        \caption{}
        \label{fig:synthetic00_tau_acc}
    \end{subfigure}
        \begin{subfigure}{0.49\columnwidth}
        \includegraphics[width = 1\columnwidth]{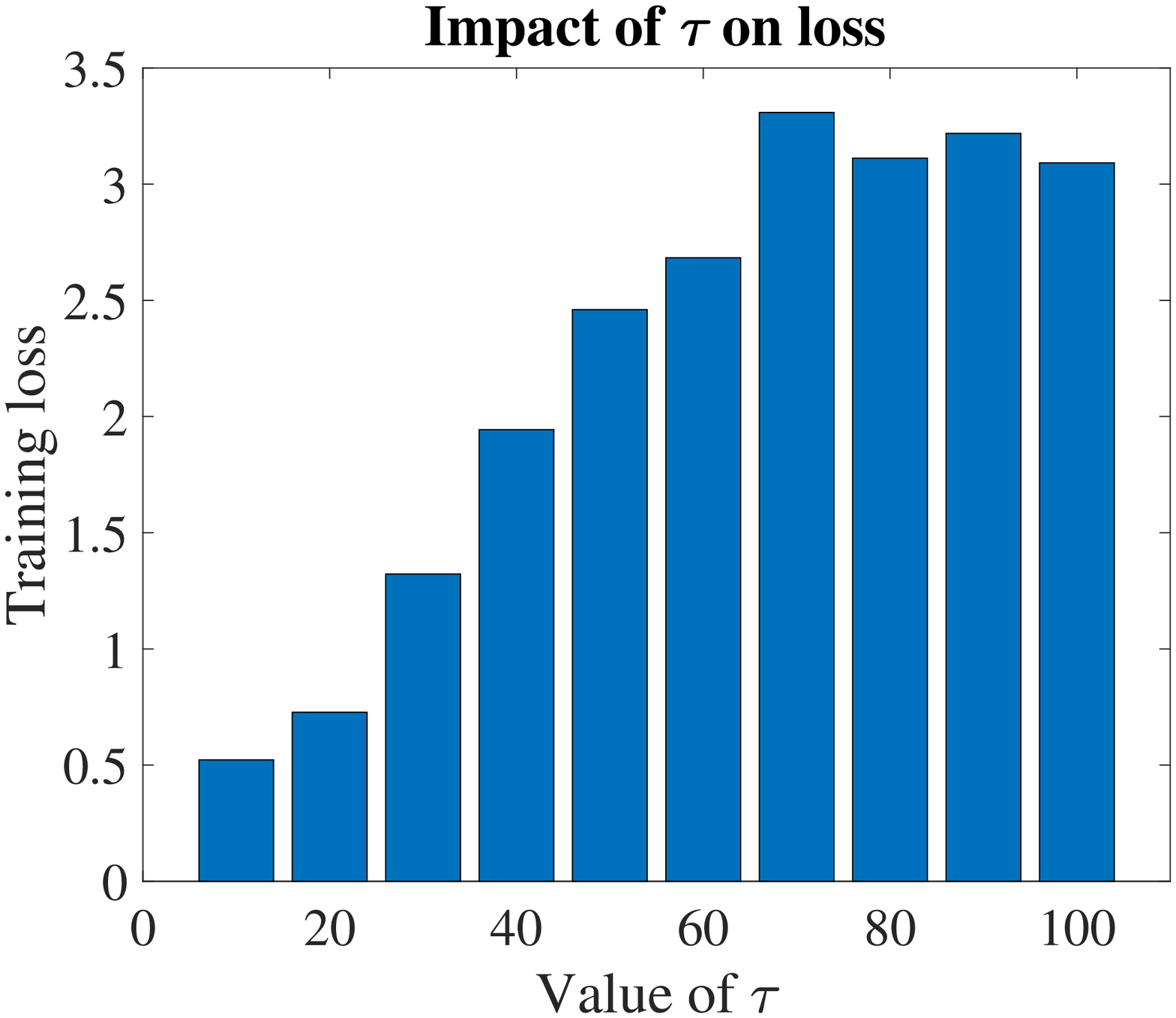}
        \caption{}
        \label{fig:synthetic00_tau_loss}
    \end{subfigure}
    \caption{Choice of $\tau$ on the non-i.i.d distributed $Synthetic(0,0)$ dataset: a) the learning accuracy of HFL with different maximum $\tau$ values; b) the training loss of HFL with different maximum $\tau$ values.}
    \label{fig:tau_00}
\end{figure}

\begin{figure}[t!]
    \centering
        \begin{subfigure}{0.49\columnwidth}
        \includegraphics[width = 1\columnwidth]{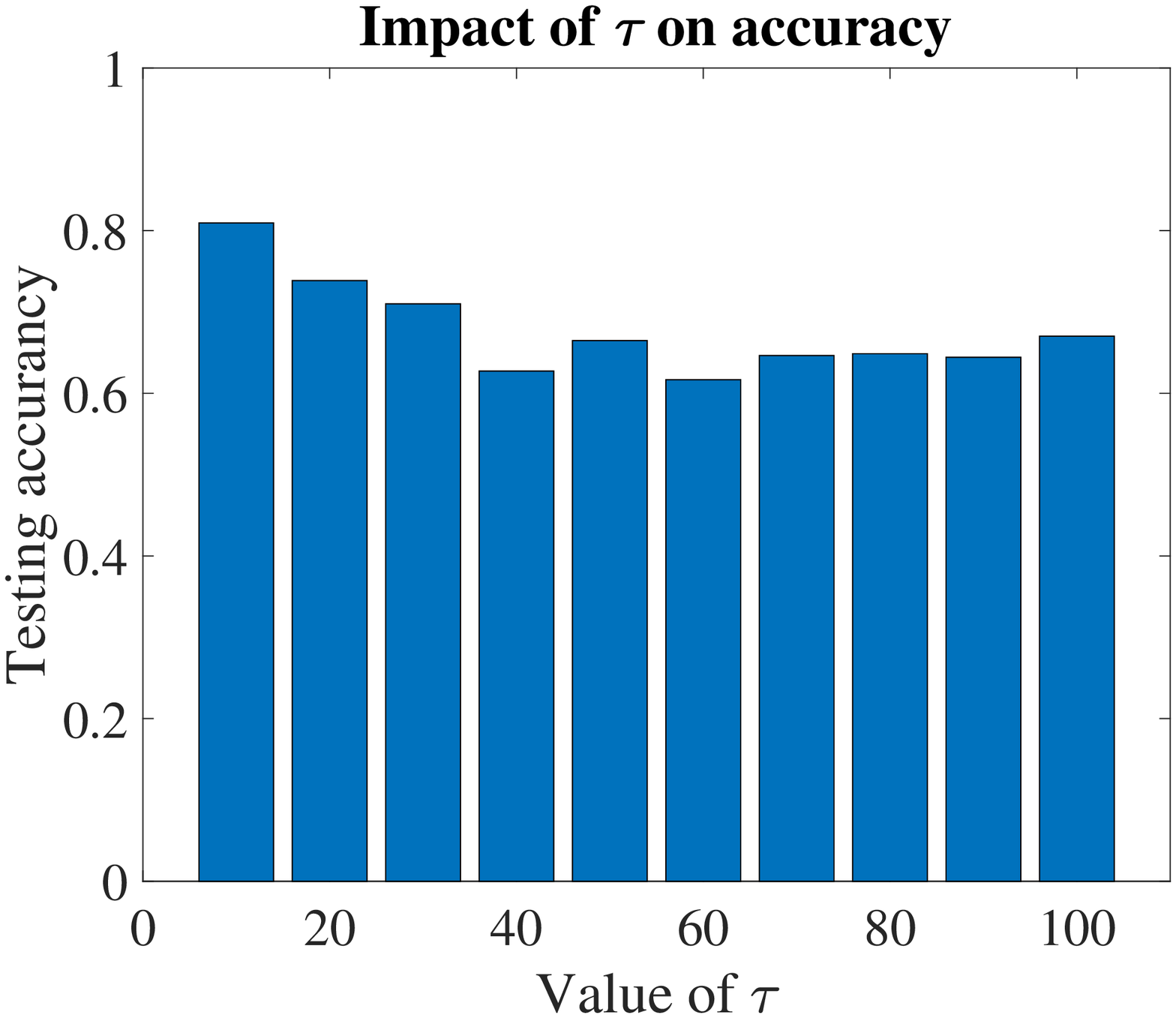}
        \caption{}
        \label{fig:synthetic00_iid_tau_acc}
    \end{subfigure}
        \begin{subfigure}{0.49\columnwidth}
        \includegraphics[width = 1\columnwidth]{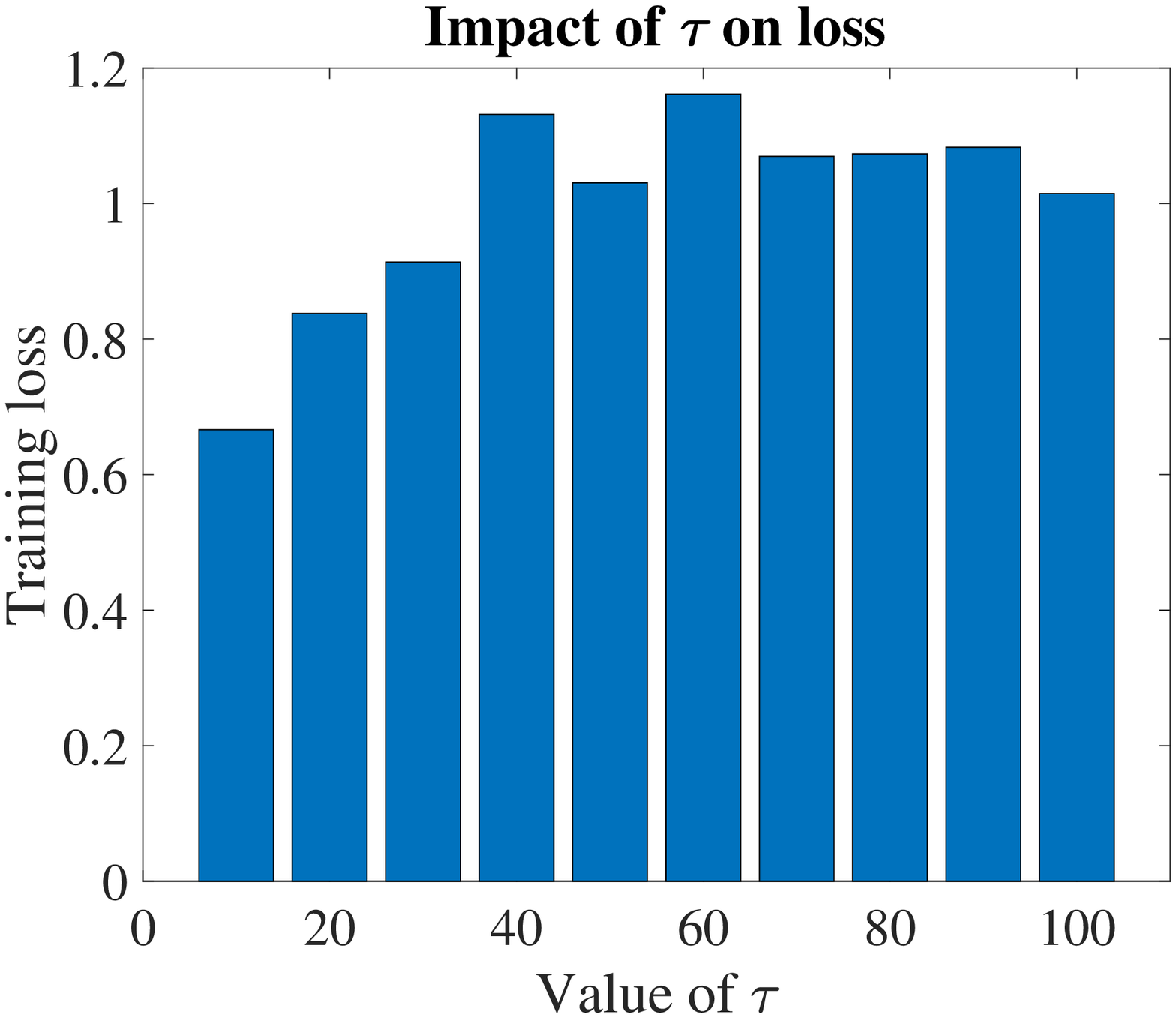}
        \caption{}
        \label{fig:synthetic00_iid_tau_loss}
    \end{subfigure}
    \caption{Choice of $\tau$ on the i.i.d distributed $Synthetic(0,0)$ dataset: a) the learning accuracy of HFL with different maximum $\tau$ values; b) the training loss of HFL with different maximum $\tau$ values.}
    \label{fig:tau_00_iid}
\end{figure}

\begin{figure}[t!]
    \centering
        \begin{subfigure}{0.49\columnwidth}
        \includegraphics[width = 1\columnwidth]{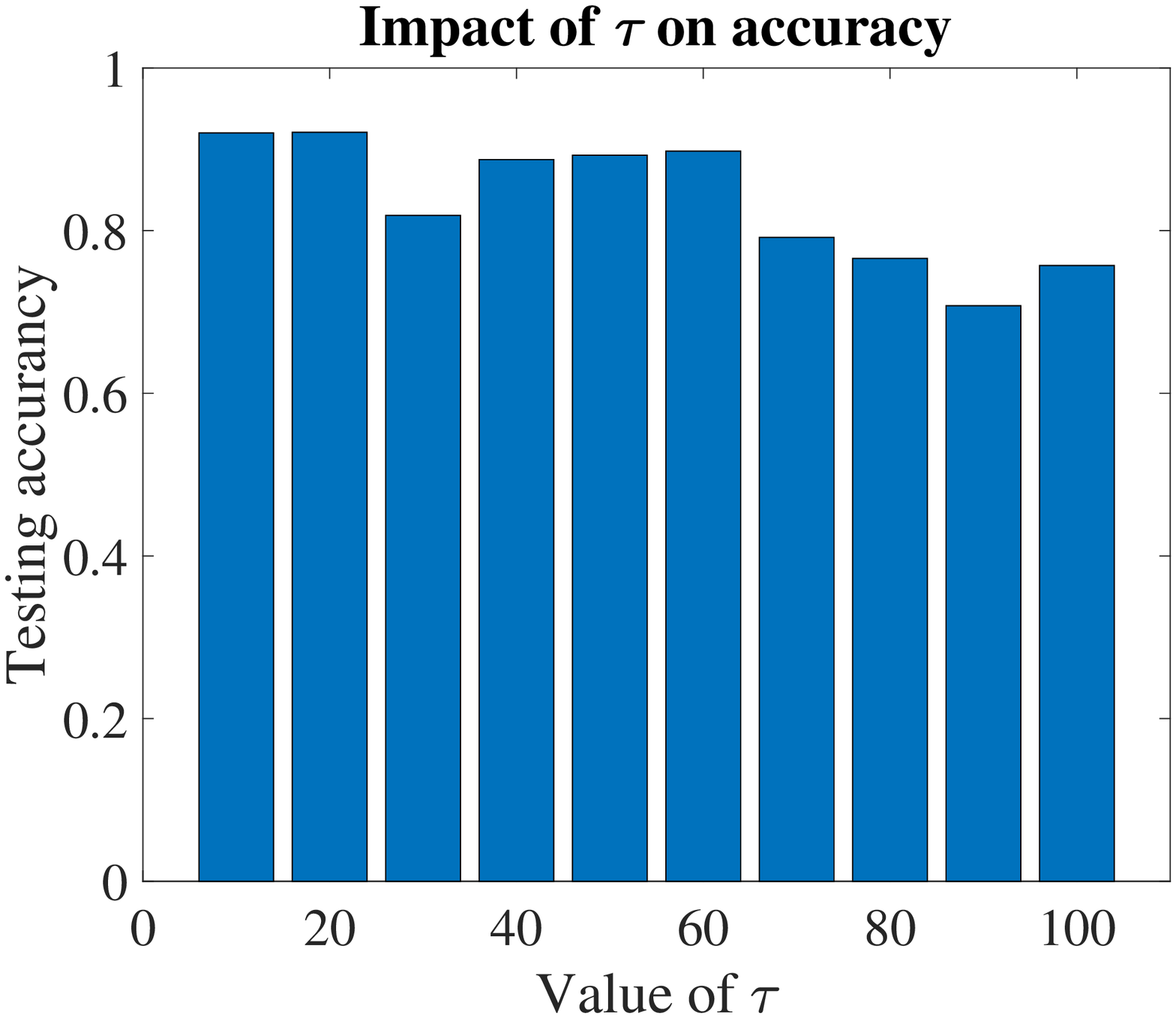}
        \caption{}
        \label{fig:tau_mnist_acc}
    \end{subfigure}
        \begin{subfigure}{0.49\columnwidth}
        \includegraphics[width = 1\columnwidth]{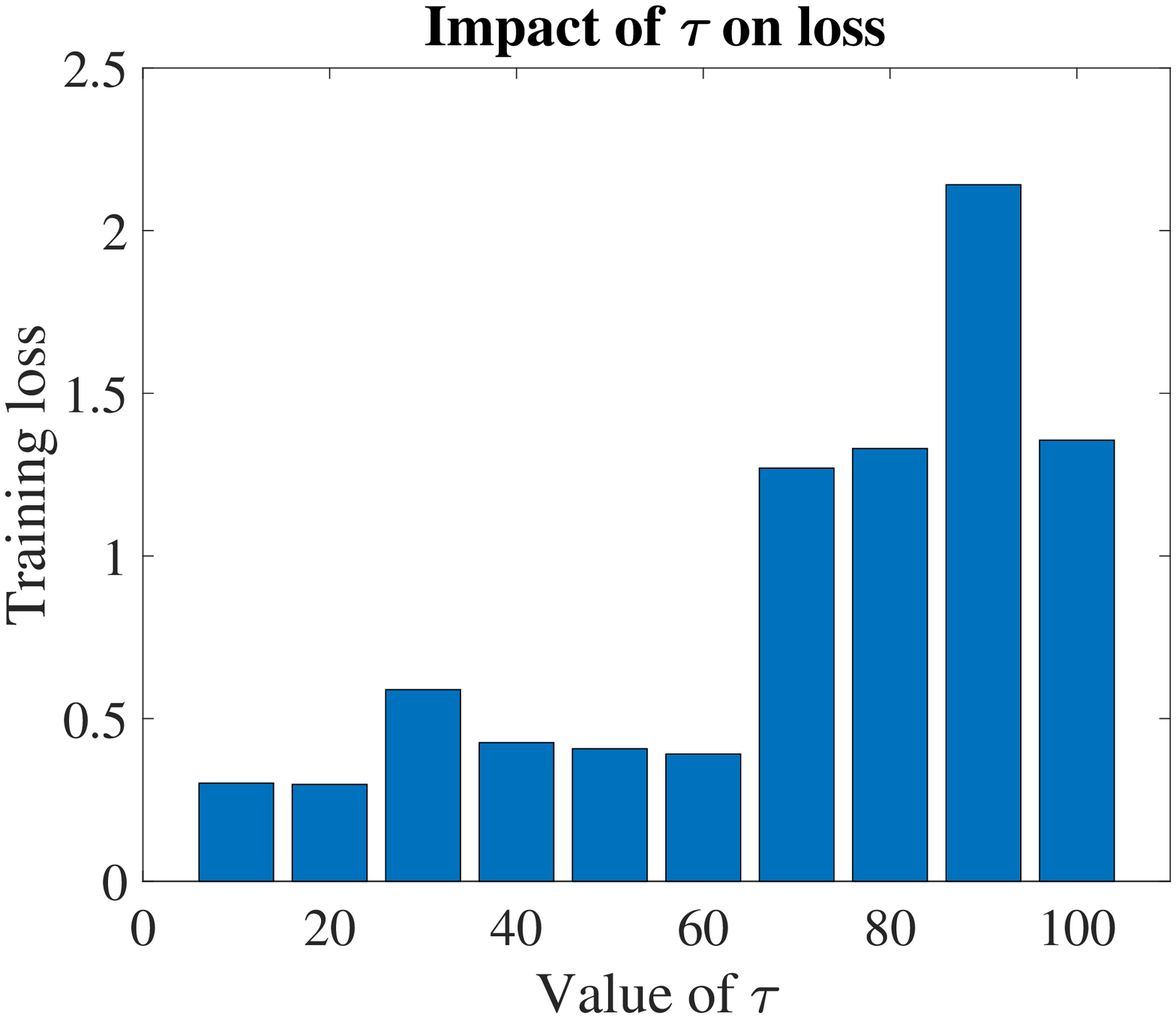}
        \caption{}
        \label{fig:tau_mnist_loss}
    \end{subfigure}
    \caption{Choice of $\tau$ on the non-i.i.d distributed MNIST dataset: a) the learning accuracy of HFL with different maximum $\tau$ values; b) the training loss of HFL with different maximum $\tau$ values.}
    \label{fig:tau_mnist}
\end{figure}

\begin{figure}[t!]
    \centering
        \begin{subfigure}{0.49\columnwidth}
        \includegraphics[width = 1\columnwidth]{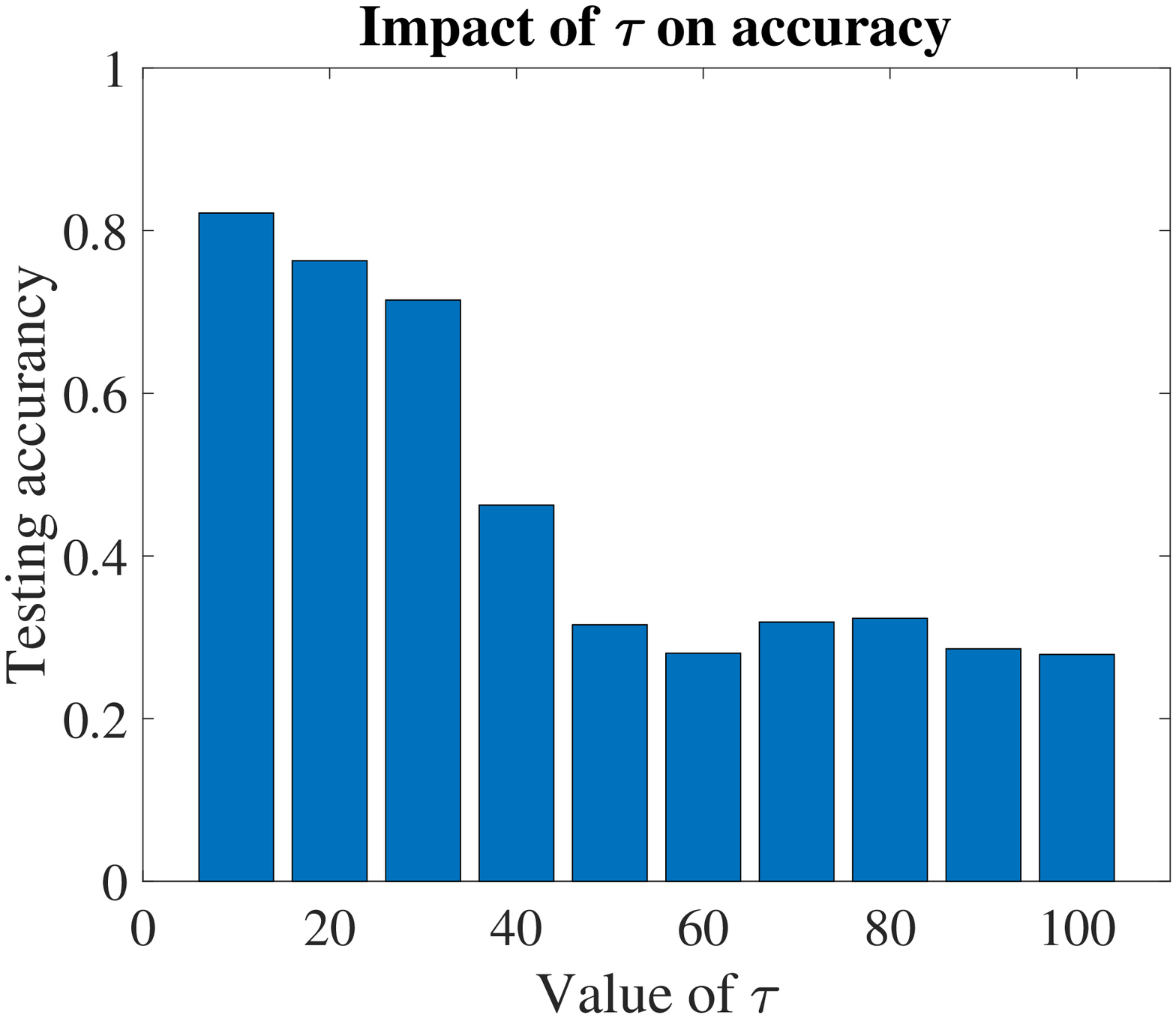}
        \caption{}
        \label{fig:tau_synthetic0505_acc}
    \end{subfigure}
        \begin{subfigure}{0.49\columnwidth}
        \includegraphics[width = 1\columnwidth]{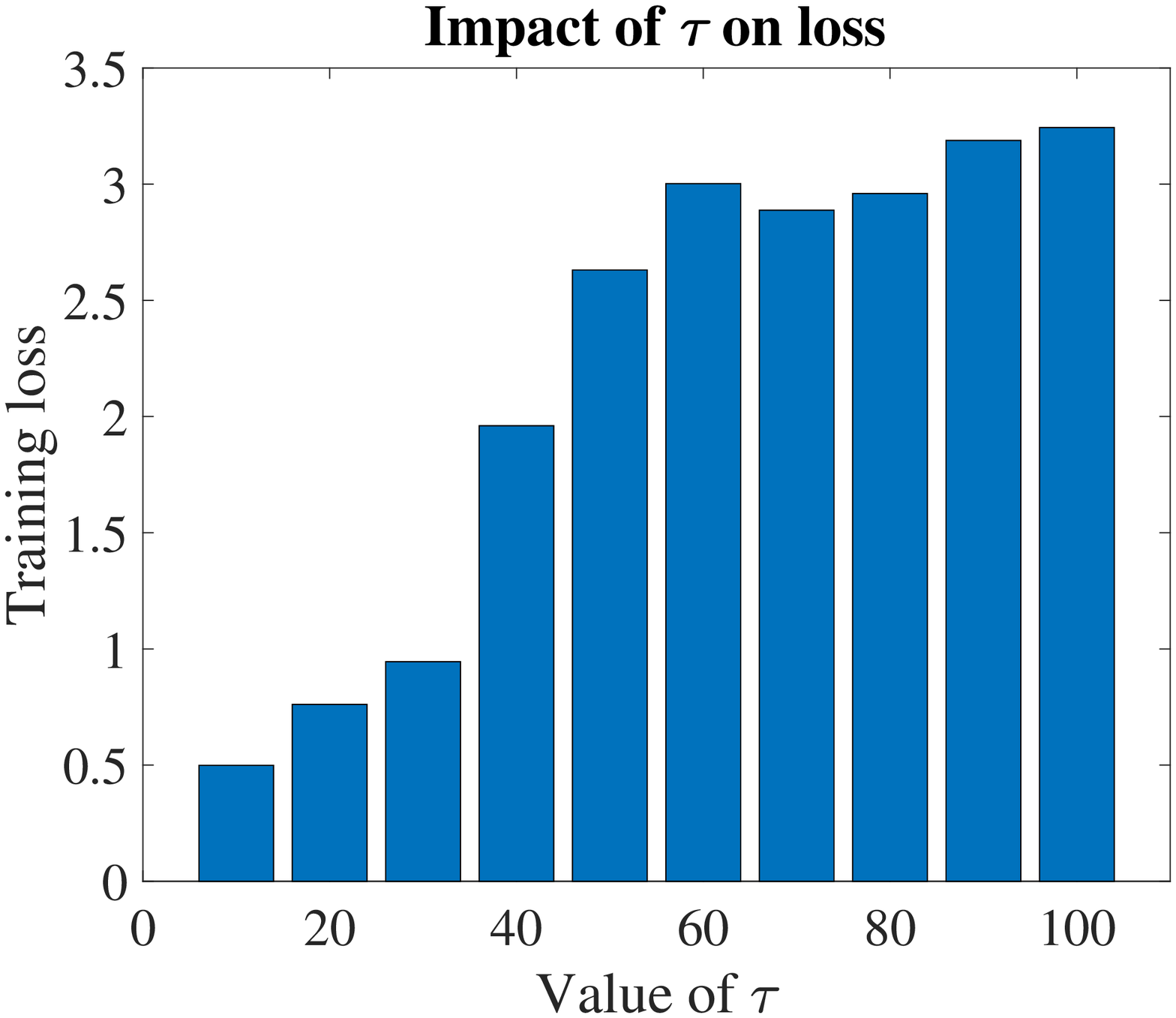}
        \caption{}
        \label{fig:tau_synthetic0505_loss}
    \end{subfigure}
    \caption{Choice of $\tau$ on the non-i.i.d distributed $Synthetic(0.5,0.5)$ dataset: a) the learning accuracy of HFL with different maximum $\tau$ values; b) the training loss of HFL with different maximum $\tau$ values.}
    \label{fig:tau_synthetic0505}
\end{figure}

\textbf{Choice of $\tau$.}
We extend the investigation of the choice of $\tau$ in the HFL algorithm. In the main paper, we discuss the choice of $\tau$ with a non-i.i.d distributed Fashion MNIST dataset. Thus, in the supplementary, we conduct multiple experiments to investigate the choice of $\tau$. For the i.i.d distributed $Synthetic(0,0)$, we introduce the choice of $\tau$ in Figure.~\ref{fig:tau_00_iid}. And for the non-i.i.d distributed $Synthetic(0,0)$, $Synthetic(0.5,0.5)$ and $Synthetic(1,1)$ datasets, we introduce the results in Figure.~\ref{fig:tau_00}-\ref{fig:tau_synthetic0505} and Figure.~\ref{fig:tau_11}. And for the MNIST dataset, we introduce the results in Figure.~\ref{fig:tau_mnist}. We could notice from those results that they supports the analysis for the choice of $\tau$ in the main paper.

\bibliography{references}
\bibliographystyle{icml2021} 

\end{document}